\documentclass[UTF8]{article}
\usepackage[T1]{fontenc}
\usepackage[utf8]{inputenc}
\usepackage{authblk}
\usepackage{amsmath}
\usepackage{amsfonts}

\usepackage{textcomp}
\usepackage{graphicx}
\usepackage{url}
\usepackage{subfigure} 
\usepackage{float} 
\usepackage{subfigure} 
\usepackage{algorithm}
\usepackage{algorithmic}
\usepackage{tikz,xcolor,hyperref}
\definecolor{lime}{HTML}{A6CE39}
\DeclareRobustCommand{\orcidicon}{
	\begin{tikzpicture}
	\draw[lime, fill=lime] (0,0) 
	circle [radius=0.16] 
	node[white] {{\fontfamily{qag}\selectfont \tiny ID}};
	\draw[white, fill=white] (-0.0625,0.095) 
	circle [radius=0.007];
	\end{tikzpicture}
	\hspace{-2mm}
}
\foreach \x in {A, ..., Z}{\expandafter\xdef\csname orcid\x\endcsname{\noexpand\href{https://orcid.org/\csname orcidauthor\x\endcsname}
			{\noexpand\orcidicon}}
}

\usetikzlibrary{graphs}
\usetikzlibrary{shapes.geometric,fit,positioning,arrows,automata,calc}
\tikzset{
  main/.style={circle, minimum size = 5mm, thick, draw =black!80, node distance = 10mm},
  connect/.style={-latex, thick},
  box/.style={rectangle, draw=black!100}
}

\title{Dive into  Decision Trees and Forests: A Theoretical Demonstration}
\author{Zhang Jinxiong \orcidA{}\\
jinxiongzhang@qq.com}
\date{} 
\providecommand{\keywords}[1]{\textbf{\textit{Keywords---}} #1}

\begin{document}

\maketitle

\begin{abstract}

Based on decision trees, many fields have arguably made tremendous progress in recent years. 
In simple words, decision trees use the strategy of ``divide-and-conquer'' 
to divide the complex problem on the dependency between input features and labels into smaller ones.
While decision trees have a long history, recent advances have greatly improved their performance in 
computational advertising, recommender system, information retrieval, etc. 

We introduce common tree-based models (e.g., Bayesian CART, Bayesian regression splines) 
and training techniques (e.g., mixed integer programming, alternating optimization, gradient descent). 
Along the way, we highlight probabilistic characteristics of tree-based models and explain their practical and theoretical benefits.
Except machine learning and data mining, we try to show theoretical advances on tree-based models from other fields such as statistics and operation research.
We list the reproducible resource at the end of each method. 

\end{abstract}


\tableofcontents

\section{Introduction}

Supervised learning is to find optimal prediction when some observed samples are given.
It is based on the belief that we can imply the global relationship of $(x, y)$ 
based on a finite partial samples $\{(x_i, y_i)\}_{i=1}^{n}$ where $n<\infty$.
In another word, it is to learn a function $f:\mathcal{X}\to\mathcal{Y}$ 
from the data set $\{(x_i, y_i)\mid x_i\in\mathcal{X}, y\in\mathcal{Y}\}_{i=1}^{n}$.
Usually  $\mathcal{X}$ is  high dimensional and it is hardly to find joint distribution of $(x,y)\in\mathcal{X\times Y}$. 
Unsupervised learning is to find some inherent structure of the data.
These learning categories cover most tasks in machine learning. 

Decision tree is a typical example of `divide-and-conquer' strategy 
and  widely used in many fields as shown 
in \cite{holzinger2015data, Maindonald2011Recursive, lakshminarayanan2016decision, zhang2003cell}.
Based on decision trees, there are improvements in diverse fields 
such as computational advertising\cite{he2014practical}, recommender system\cite{liu2015boosting}, 
information retrieval\cite{lucchese2017quickscorer, hu2019unbiased, dato2016fast}.
It is demonstrated the effectiveness of recursive partitioning specially 
for the tasks in health sciences in \cite{Maindonald2011Recursive}.
In \cite{zhou2017deep}, it is to solve image recognition task based on the ensemble of decision trees.
Although decision trees are usually applied to regression and classification tasks, 
it is available to apply tree-based methods to unsupervised tasks such as in 
\cite{leoII, criminisi2012decision, liu2008isolation, strobl2007bias}.
There is a cruated list of decision tree research papers 
in \UrlBigBreaks{https://github.com/benedekrozemberczki/awesome-decision-tree-papers}.

There is a chronological literature review in \cite{criminisi2013decision}, which summarizes the important work of decision trees in last century.
In \cite{criminisi2012decision} applies tree-based methods to diverse tasks such as classification, regression, density estimation and manifold learning.
The \cite{holzinger2015data} covers many aspects of  very important technique in decision tree  from the data mining viewpoint.
Here we will focus on the recent advances on this fields.

It is usually represented in a tree structure visually.
\begin{figure}[H] 
	\centering 
	\includegraphics[width=0.85\textwidth]{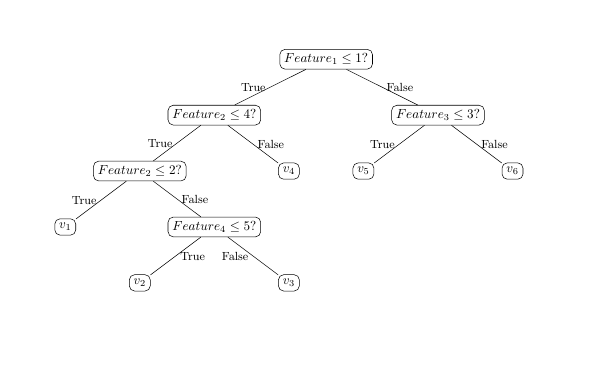}
	\caption{A toy example of decision tree}  
	\label{Fig.Toy} 
\end{figure}
Conventionally, visual representation of tree is in the top-to-down order as shown in \eqref{Fig.Toy}. 
And decision trees share many terms with tree data structure.
For example, the topmost node is called root node; 
if the node $a$ is directly under the node $b$, we say the node $a$ is a child of the node $b$ and
the $b$ is the parent of node $a$;
the nodes without any children are called leaves or terminal nodes.

The decision tree is formally expressed as the simple function
\begin{equation}\label{Decision Tree}
T(x)=\sum_{j}^{J}\gamma_j\mathbb{I}(x\in R_j)
\end{equation}
with parameters $\{\gamma_j, R_j\}_{j=1}^{J}$.
The number of leaf nodes $J$ is usually treated as a hyper-parameter.
And the set family $\{R_j\}_{j=1}^{J}$ is a partitioning of the domain of $x$.
Each set $R_j$ is parameterized with $\theta_j$.
The indicator function $\mathbb{I}(x\in R_i)$ is the characteristic function of the set  $R_i$,defined as
$$\mathbb{I}(x\in R_i)=\begin{cases}
1, &\text{if }{x\in R_i};\\
0, &\text{otherwise}.
\end{cases}$$

Random forest is the sum of decision trees with some random factors.
The trees in random forest not only draw some samples from the train set but also take a part of the sample features during tree induction.
Boosting decision trees method is to add up some well-designed decision trees with appreciate weights
$$\sum_{n}^{N}w_nT_n(x).$$
Some efficient implementation of this boosting are open and free such as \cite{chen2015xgboost, prokhorenkova2017catboost:, ke2017lightgbm:}.
The weight $w_n$ depends on the error of $T_{n-1}$ for $n\geq 2$ thus the trees in such methods are trained in a sequential way.

Both above methods are additive trees as application of ensemble methods to decision trees.
With the help of Bayesian methods, we can invent the `multiplicative' trees in some sense.
And not only one term of the sum in $T(x)$ can be nonzero (associated with the leaf in which $x$ falls at random) in Bayesian trees.
We will revisit methods of overcoming the deficiencies of classic decision trees under the probabilistic len.

Additionally, we will revisit the relations of tree-baed models,  Bayesian hierarchical models and deep neural networks.
And from a numerical optimization  perspective, we will review non-greedy optimization methods for decision trees.

We would review the following the topics to understand the properties of decision trees:
\begin{enumerate}
 \item Decision trees: basic introduction to decision trees and related fields.
 \item Additive decision trees: linear aggregation of decision trees. 
 \item Probabilistic decision trees: probabilistic thoughts and methods in decision trees.
 \item Bayesian decision trees: Bayesian ideas in tree-based models.
 \item Optimal decision trees: non-greedy induction methods of decision trees.
 \item Neural decision trees: the combination of decision trees and deep neural networks.
 \item Regularized decision trees: regularization in tree-based models.
 \item Conditional computation: extension of tree-based models.
\end{enumerate} 

\section{Decision trees}


\subsection{Algorithmic  construction for decision trees}

Decision tree is generally regarded as a stepwise procedure, consisting of two phases- the induction or growth phase and  the pruning phase.
Both phases are of the `if-then' sentences so it is possible to deal with the mixed types of data.

During the induction or growth of decision tree, it is first to recursively  partition the domain of input space.
In another word, it is to find the subset $R_j$ for $j=1,2, \cdots, J$ in \eqref{Decision Tree}.
The second step is to determine the local optimal model at each subset.
In another word, it is to find the best $\gamma_j$ according to some criteria in \eqref{Decision Tree}.
Majority of the variants of decision tree is to improve the first phase.
We quote the pseudocode for tree construction in \cite{loh2011classification}.

\begin{algorithm}[h]\label{tree construction}
	\caption{Pseudocode for tree construction by exhaustive search}
	\begin{algorithmic}[1]
		\STATE {Start at the root node}
		\STATE {For each $X$, find the set $S$ that minimizes the sum of the node impurities in the two child nodes and choose the split $X^{*} \in S^{*}$
			that gives the minimum overall $X$ and $S$.}
		\STATE If a stopping criterion is reached, exit. Otherwise, apply step 2 to each child node in turn.
	\end{algorithmic}
\end{algorithm}
Different node impurities and stopping criterion lead to different tree construction methods.
Here we would not introduce the split creteria such as entropy or the stopping criterion.

To avoid the over-fitting and improve the generalization performance, pruning is to decrease the size of the decision tree.
In \cite{fridedman1991multivariate}, there is the differences between the backwards variable selection and pruning.

Decision tree is an adaptive computation method for function approximation.
However the framework\cite{loh2011classification} does not minimize the cost directly via numerical optimization methods such as gradient descent because it only minimizes the sum of the node impurities at each split.
Mixed integer programming is used to find the optimal decision trees such as 
\cite{bertsimas2007classification, bertsimas2017optimal, DBLP:conf/bnaic/VerhaegheNPQS19, DBLP:journals/corr/MenickellyGKS16, bennett1996optimal}, 
which is discussed later. 

We pay attention into the modification of this framework and the connection between decision trees and other models.

\subsection{Representation of decision tree}
Here we focus on the representation of decision tree.
In another word, we pay attention to inference of a trained decision tree rather than the traning of a decision tree given some dataset.

Decision trees are named because of its graphical representation where every input travels 
from the top (as known as root) to the terminal node (as known as leaf).
It is the  recursive ‘divide-and-conquer’ nature which makes it different from other supervised learning methods.
It is implemented as a list of `IF-THEN' clauses. 

In \cite{fridedman1991multivariate}, the recursive partitioning regression, as binary regression tree,  
is  viewed in a more conventional light as a stepwise regression procedure:
\begin{equation}\label{sum-product}
f_M(x)=\sum_{m=1}^{M}a_m\prod_{k=1}^{K_m}H[s_{km}(x_{v(k, m) - t_{km}})]
\end{equation}
where $H(\cdot)$ is the unit step function.
The quantity $K_m$, is the number of splits that gave rise to basis function.
The quantities $s_{km}$ take on values k1and indicate the (right/left) sense of the associated step function.
The $v(k, m )$ label the predictor variables and 
the $t_{km}$, represent values on the corresponding variables.
The internal nodes of the binary tree represent the step functions and the terminal nodes represent the final basis functions.
It is hierarchical model using a set of basis functions and stepwise selection.
The product term $\prod_{k=1}^{K_m}H[s_{km}(x_{v(k, m) - t_{km}})]$ is an explicit 
form of the indicator function $\mathbb{I}$ in \eqref{Decision Tree}.  

Yosshua Bengio, Olivier Delalleau, and Clarence Simard in \cite{bengio2010decision}
define the decision tree $T:\mathbb{R}^d \to \mathbb{R}$ as an additive model as below:
\begin{equation}\label{DTree}
T(x)=\sum_{i\in leaves}g_i(x)\mathbb{I}_{x\in R_i}=\sum_{i\in leaves}g_i(x)\prod_{a\in ancestors(i)}\mathbb{I}_{S_a(x)=c_{a, i}}
\end{equation}
where ${R}_i \subset \mathbb{R}^d$ is the region associated with leaf $i$ of the tree, 
$ancestor s(i)$ is the set of ancestors of leaf node $i$, 
$c_{a,i}$ is the child of node a on the path from a to leaf $i$, 
and $S_a$ is the $n$-ary split function at node $a$. 
The indicator function $\mathbb{I}_{x\in R_i}$ is equal to $1$ if $x$ belongs to $R_i$ otherwise it is equal to $0$.
$g_i (\cdot)$ is the prediction function associated with leaf $i$ and
is learned according to the training samples in $R_i$. 

The prediction function $g_i(x)$ is always restricted in a specific function family such as polynomials \cite{chaudhuri1994piecewise}.
Usually, $g_i$ is constant when we call \eqref{DTree} classical decision trees.
For example, decision tree in XGBoost\cite{chen2015xgboost}  is expressed as 
\begin{equation}\label{chen}
T(x) = w_{q(x)}, w\in R^{T}, q:R^d\to \{1,2,\cdots, T\}
\end{equation}
where $w$ is the vector of scores on leaves, $w_i$ is the score on leaf $i$ for $\forall i \in\{1,2,\cdots, L\}$;
$q$ is a function assigning each data point to the corresponding leaf, and $T$ is the number of leaves.

In \cite{kuralenok2019monoforest},  each tree is transformed into a well-known polynomial form.
It is to express the decision tree in algebraic form
\begin{equation}\label{Mono}
h(x)=\sum_{\ell\in leaves}w_{\ell}\mathbb{I}\{x\in \ell\}
\end{equation}
where the indicator function is a product of indicators induced by splits along the path from root to the terminal node:
$$\mathbb{I}\{x\in \ell\}=\prod_{c\in\text{right splits}}c(x)\prod_{c\in\text{left splits}}(1-c(x)).$$

The key components of decision trees are 
(1) the split functions to partition the sample space at each leaf such as the split function $S_a$ in \eqref{DTree};
(2) the prediction of each leaf or terminal node such as the decision function $g_i$ in \eqref{DTree}.
Usually each split function is univariate thus decision tree is inherently sparse. 
Note that exactly one term of the sum in $T(x)$ \eqref{DTree} can be nonzero (associated with the leaf in which $x$ falls). 
In the term of mathematics, decision tree is exactly the simple function when $g_i(x)$ is constant forall $i$.
The training methods are to find proper $g_i(x)$ and $R_i$.
Diverse tree-based methods are to modify the training methods  as shown in \cite{loh2014fifty, loh2011classification, holzinger2015data, Maindonald2011Recursive}.

\section{Additive decision trees}

Additive decision trees as well as its variant is to overcome the following fundamental limitation of decision tree via emsemble methods
\begin{itemize}
	\item  the lack of continuity,
	\item  the lack of smooth decision boundary,
	\item  the high instability with respect to minor perturbations in the training data,
	\item  the inability to provide good approximations to certain classes of simple often-occurring functions.
\end{itemize}

We call the sum of decision trees as additive decision tree defined in the following form
\begin{equation}\label{Additive DT}\sum_{n=1}^{N}w_n T_{n}\end{equation}
where $T_n$ is a decision tree and $w_n$ is greater than $0$ for $n=1,2,\cdots, N$.
For convenience,  we divide the additive decision trees  into two categories:
(1) decision forests such as \cite{criminisi2012decision, liu2008isolation, zhang2003cell};
(2) boosting trees such as \cite{chen2015xgboost, prokhorenkova2017catboost:, ke2017lightgbm:, luna2019building}.

\begin{figure}[H]
  \centering 
  \includegraphics[width=0.85\textwidth]{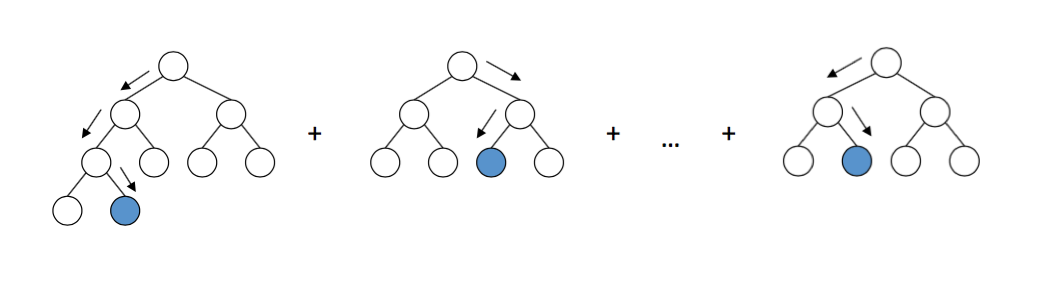}
  \caption{Additive decision trees}  
  \label{ADT} 
\end{figure}

The main difference between them are the methods to train the single decision tree and the weights to combine them.
Following Leo Breiman in \cite{breiman1996bias}, the decision forest is 
to resample from the original training set to construct a single decision tree and and then combined by voting;
gradient boosting trees are  to adaptively resample and combine (hence the acronym--arcing) 
so that the weights in the resampling are increased for those cases most often misclassified and 
the combining is done by weighted voting.
Leo Breiman gave his Wald Lecture on this topic \cite{leoI, leoII}.
Both methods are explained as interpolating classifiers in \cite{wyner2017explaining}.


\subsection{Decision forest}

A decision forest is a collection of decision trees by a simple averaging operation \cite{breiman2001random}.
Antonio Criminisi and his coauthors wrote two books on this topic \cite{criminisi2012decision, criminisi2013decision}.
Forest-based classification and prediction is also discussed in the Chapter 6 in \cite{Maindonald2011Recursive}.
All trees in the decision forest are trained independently (and possibly in parallel). 
Decision forests are designed to improve the stability of a single decision tree.

And random forest will permute the training set and combine the trained trees.
Different methods on the permutation of the training data leads to different random forests.
The name `random forest` is initially  proposed in \cite{breiman2001random}.
Here we refer it to additive trees \eqref{Additive DT} which use the  permuted  data sets to train the component tree.
For examples,  
Leo Breiman in \cite{breiman2001random} use a bootstrap samples and truncate the feature space on random to train each tree 
in order to decrease the correlation between the trees, 
which can be regarded as a combination of \href{https://en.wikipedia.org/wiki/Bootstrap_aggregating}{bagging} 
and \href{https://en.wikipedia.org/wiki/Random_subspace_method}{random subspace methods}.

In an extremely randomized trees \cite{geurts2006extremely}  this is made much faster by the following procedure.
\begin{enumerate}
	\item The feature indices of the candidate splits are determined by drawing \textbf{max\_features} at random.
	\item Then for each of these feature index, we select the split threshold by drawing uniformly between the bounds of that feature.
\end{enumerate}

When the set of candidate splits are obtained, as before we just return that split that minimizes $C$.


Another direction in this field is to decrease the number of decision trees with low accuracy reduction.
It is the model compression for forest-based models.
For example, it is shown that ensembles generated by a selective ensemble algorithm, 
which selects some of the trained C4.5 decision trees to make up an ensemble, 
may be not only smaller in the size but also stronger in the generalization than ensembles generated by non-selective algorithms\cite{zhou2003selective}.
Heping Zhang and Minghui Wang proposed a specific method  to find a sub-forest (e.g., in a single digit number of trees) 
 that can achieve the prediction accuracy of a large random forest (in the order of thousands of trees) in \cite{zhang2009search}.

Yi Lin and Yongho Jeon  study random forests through their connection with a new framework of
adaptive nearest neighbor methods \cite{lin2006random}.



\subsection{Boosted decision trees}

Breiman in \cite{breiman1998arcing} summarized that the basic idea of AdaBoost \cite{freund1995desicion} is to adaptively resample and combine (hence the acronym--arcing) 
so that the weights in the resampling are increased for those cases most often misclassified and the combining is done by weighted voting.
Then in \cite{breiman1997arcing} it is found that arcing algorithms  are optimization algorithms which minimize some function of the edge.
A general gradient descent ``boosting” paradigm is developed for additive expansions based on any fitting criterion in \cite{friedman2001greedy}
There are diverse algorithms and theories on boosting as shown by Schapir and Freund in the monograph \cite{schapire2012boosting}.

Multiple Additive Regression Trees (MART\textsuperscript{TM}) is an implementation of the gradient tree boosting methods as described in \cite{friedman2001greedy}
Although there are other boosting methods as discussed in \cite{leoI}, 
gradient boosting decision trees are more popular than others such as the open and free softwares 
\cite{chen2015xgboost, prokhorenkova2017catboost:, ke2017lightgbm:, wen2018efficient}.

\begin{algorithm}[h]
\caption{Gradient Boost Decision Trees}
\begin{algorithmic}[1]
\STATE {Input training data set $\{(x_n, y_n)\mid x_n\in\mathrm{X}\subset\mathbb{R}^{p}, y_n\in\mathbb{R},n=1,2,\cdots, N\}$}
\STATE{ Initialize $f_0=\arg\min_{\gamma}\sum_{i=1}^{N}L(x_i, \gamma)$}
\FOR{$t = 1, 2, \dots, T$}
\FOR{$i = 1, 2, \dots , N$}
\STATE Compute $r_{i,t}=-{[\frac{\partial L(\mathrm{y}_i, f(x_i))}{\partial f(x_i)}\mid_{f=F^{(t-1)}}]}.$
\ENDFOR
\STATE  Fit a regression tree to the targets $r_{i,t}$   giving terminal regions
   $$R_{j,m}, j = 1, 2,\dots , J_m. $$
\FOR{$j = 1, 2,\dots , J_m$ }
\STATE  Compute $\gamma_{j,t}=\arg\min_{\gamma}\sum_{x_i\in R_{j,m}}{L(\mathrm{y}_i, F^{(t-1)}(x_i)+\gamma)}$.
\ENDFOR
\STATE  $f_t={\sum}_{j=1}^{J_m}{\gamma}_{j, t} \mathbb{I}(x\in R_{j, m})$
\STATE Update $F^{(t)} = F^{(t-1)}+\nu f_t,\nu\in(0, 1)$
\ENDFOR
\STATE Output $F^{(T)}$.
\end{algorithmic}
\end{algorithm}


\section{Probabilistic decision trees and forest}

We begin with \href{https://www.sciencedirect.com/science/article/pii/B9780080510552500110#!}{John Ross Quinlan}:
\begin{quote}
	Decision trees are a widely known formalism for expressing classification knowledge 
	and yet their straightforward use can be criticized on several grounds. 
	Because results are categorical, they do not convey potential uncertainties in classification. 
	Small changes in the attribute values of a case being classified may result in sudden and inappropriate changes to the assigned class. 
	Missing or imprecise information may apparently prevent a case from being classified at all. 
\end{quote}

As argued in \cite{lakshminarayanan2016decision}, the probabilistic approach allows us to encode prior assumptions about tree structures 
and share statistical strength between node parameters; 
furthermore, it offers a principled mechanism to obtain probabilistic predictions
which is crucial for applications where uncertainty quantification is important.
We will revisit methods of overcoming these deficiencies under the probabilistic len such as
\cite{lakshminarayanan2016decision, 10.1145/3022671.2984041}.
Another motivation of probabilistic trees is to soften their decision boundaries as well as the fuzzy decision trees.

Let us review the sum-product representation of decision tree
\begin{equation}\nonumber
f(x)=\sum_{m=1}^{M}a_m\prod_{k=1}^{K_m}H[s_{km}(x_{v(k, m) - t_{km}})]
\end{equation}
where $H[\cdot]\in\{0, 1\}$ which can be regarded as binary probability distribution.
The product $\prod_{k=1}^{K_m}H[s_{km}(x_{v(k, m) - t_{km}})]$ is the probability reaching the leaf node.
In short, soft decision tree replace the indicator function with with continuous functions in $[0, 1]$ such as cumulant density function or membership function.
In particular, the output  of probabilistic trees is defined as below
\begin{equation}\nonumber
f(x)=\sum_{m=1}^{M}a_m\prod_{k=1}^{K_m}H[s_{km}(x_{v(k, m) - t_{km}})]
\end{equation}
where $H(\cdot)$ is cumulant density function.

\subsection{Hierarchical mixtures of experts}
The mixtures-of-experts (ME) architecture is a mixture model in which the mixture components are conditional probability distributions.
The basic idea of Probabilistic Decision Trees (also called Hierarchical Mixtures of Experts)\cite{jordan1994hierarchical} is to convert the decision tree into a mixture model.
Probabilistic Decision Trees is a blend of Bayesian methods and artificial neural networks.
To quote Michael I Jordan \cite{Michael}
\begin{quote}
	Probabilistic Decision Trees
	\begin{itemize} 
		\item drop inputs down the tree and use probabilistic models for decisions; 
		\item at leaves of trees use probabilistic models to generate outputs from inputs;
		\item use a Bayes` rule recursion to compute posterior credit for  non-terminal nodes in the tree.
	\end{itemize}
\end{quote}
The routing and outputs of of probabilistic decision trees are probabilistic
while their structures  are  determined rather than probabilistic. 
In another word, the number and positions of nodes are determined.

The architecture is a tree in which the gating networks sit at the non-terminal nodes of the tree. 
These networks receive the vector $x$ as input 
and produce scalar outputs that are a partition of unity at each point in the input space. 
The expert networks sit at the leaves of the tree. 
Each expert produces an output vector $p_{ij}$ for each input vector. 
These output vectors proceed up the tree, being multiplied by the gating network outputs and summed at the non-terminal nodes.

\begin{table}  
	\caption{Comparison of decision trees and probabilistic decision trees}  
	\begin{center}  
		\begin{tabular}{p{5cm}|p{5cm}}
			\hline  
			Decision Trees       & Probabilistic Decision Trees \\
			\hline  
			\hline
			Test functions        & Gating networks \\
			\hline
			Prediction functions  & Expert networks \\
			\hline
			Recursively partitioning of the input space &  Soft probabilistic splits of the input space\\
			\hline
			Simple functions      & Smoothed piecewise analogs of the corresponding generalized linear models\\
			\hline
			Determined routing    & Stochastic routing \\
			\hline
		\end{tabular}  
	\end{center}
\end{table}

We follow the generative model interpretation of probabilistic decision trees described by Christopher M. Bishop and Markus Svensen in \cite{bishop2002bayesian}.
Each gating node has an associated binary variable $z_i\in\{0, 1\}$ whose value is chosen with probability given by
$$p(z_i\mid X, v_i)=\sigma(x^Tv_i)^{z_i}(1-\sigma(x^Tv_i))^{1-z_i}$$
where $\sigma(x)=\frac{1}{1+\exp(-x)}$ for $x\in\mathbb{R}$ is the logistic sigmoid function, and
$v_i$ is vector of parameters governing the distribution.
If $z_i=1$ we go down the left branch while $z_i=0$ we go down the right branch.
Starting at the top of the tree, we thereby stochastically choose a path down to a single expert node $j$,
and then generate a value for $t$ from conditional distribution for that expert.
Given the states of the gating variables, the HME model corresponds to a conditional distribution for $t$ of the form
$$p(t\mid x, W, \tau, z)=\prod_{j=1}^{m}\mathcal{N}(t\mid X, W_jx, \tau^{-1}I)^{\zeta_j}$$
where $M$ is the total number of experts, $W$ denotes $\{W_j\}$ and $\tau$ denotes $\{\tau_j\}$.
Here we have defined 
$$\zeta_j=\prod_{i}\tilde{z}_j$$
in which the product is taken over all gating nodes on the unique path from the root node to the $j$th expert, and
$$\tilde{z}=\begin{cases}z_i, &\text{if $i$ is in the left sub-tree of the node $i$}\\
1-z_i,&\text{otherwise}.\end{cases}$$ 
Marginalizing over the gating variables $z=\{z_i\}$,
\begin{equation}\begin{split}
p(t\mid x, W, \tau, v) &=\sum_{z}\prod_{j=1}^{m}\mathcal{N}(t\mid X, W_j x, \tau^{-1}I)^{\zeta_j}\prod_{i} p(z_i\mid x, v_i)\\
&= \sum_{j}{\pi}_j(x) \mathcal{N}(t\mid X, W_jx, \tau^{-1}I)
\end{split} \end{equation}
so that the conditional distribution $p(t\mid x, W, \tau, v)$ is a mixture of Gaussians in 
which the mixing coefficients ${\pi}_j(x)$ for expert $j$ is given by a product over all gating nodes on the 
unique path from the root to expert $j$ of factor $\sigma(x^Tv_i)$ or $1-\sigma(x^Tv_i)$ 
according to whether the branch at the $i$th node corresponds to $z_i=1$ or $z_i=0$.


\subsubsection{Bayesian hierarchical mixtures of experts}

Christopher M. Bishop and Markus Svensen  combine `local’ and `global’ variational methods 
to obtain a rigorous lower bound on the marginal probability of the data in \cite{bishop2002bayesian}.

We define a Gaussian prior distribution independently over of the parameters $v_i$ for each of the gating nodes given by
$$p(v_i\mid \beta_i)=\mathcal{N}(v_i, \beta^{-1}_i\mathbb{I}).$$
Similarly, for the parameters $W_j$ of $j$th expert nodes we define priors given by
$$p(W_j\mid \alpha_j)=\prod_{i}^{d}\mathcal{N}(w_{ji}, \alpha^{-1}_{j}\mathbb{I}),$$
where $j$ runs over the target variables, and $d$ is the dimensionality of the target space.
The hyperparameter $\alpha_j$, $\beta_i$ and $\tau_j$ are given conjugate gamma distributions.

The variational inference algorithm for HME is extended  by using automatic relevance determination (ARD) priors in \cite{mossavat2011sparse}.

\subsubsection{Hidden Markov decision trees}

In \cite{jordan1997hidden}, it is to combine the Hierarchical Mixtures of Experts and the hidden Markov models to produce the hidden Markov decision tree 
for time series.

This architecture can be viewed in one of two ways:
(a) as a time sequence of decision trees 
in which the decisions in a given decision tree depend probabilistically on the decisions 
in the decision tree at the preceding moment in time; 
(b) as an HMM in which the state variable at each moment in time is factorized  
and the factors are coupled vertically to form a decision tree structure.

\subsubsection{Soft decision trees}

As opposed to the hard decision node 
which redirects instances to one of its children depending on the outcome of test functions, 
a soft decision node redirects instances to all its children with probabilities calculated by a gating function \cite{irsoy2012soft}.

Learning the tree is incremental and recursive, as with the hard decision tree. The algorithm starts with one node and fits a constant model. 
Then, as long as there is improvement, it replaces the leaf by a subtree.
This involves optimizing the gating parameters and the values of its children leaf nodes by gradient-descent over an error function.

\href{https://hci.iwr.uni-heidelberg.de/content/end-end-learning-deterministic-decision-trees}{End-to-end Learning},
\href{https://pure.tudelft.nl/portal/files/66632115/Hehn2019_Article_End_to_EndLearningOfDecisionTr.pdf}{End-to-End Learning of Decision Trees and Forests},
is an Expectation Maximization training scheme for decision trees that are fully probabilistic at train time, 
but after a deterministic annealing process become deterministic at test time.

Each leaf is reached by precisely one unique set of split outcomes, called a path. 
We define the probability that a sample $x$ takes the path to leaf $\ell$ as
$$\mu_{\ell}(x)=\prod_{n\in \mathcal{A}_L(\ell)}s(f_n)\prod_{n\in \mathcal{A}_R(\ell)}(1-s(f_n))$$
where $f_n$ is the split/test/decision function of the non-terminal node $n$ 
such as a linear combination of the input $f_n(x)=\left<s_n,x\right>-b_n$;
$s$ is the probability density function such as $s(x)=\frac{1}{1+\exp(-x)}$;
$A_L(\ell)$($\mathcal{A}_R(\ell)$) is the set of ancestors of $\ell$ 
  whose left(right) branch has been followed on the path from the root node to $\ell$.
The prediction of the entire decision tree is given by multiplying the path probability with the corresponding leaf prediction
$$p(y)=\sum_{\ell\in\mathcal{T}_{L}}(\pi_{\ell})_y\mu_{\ell}(x)$$
where $\pi_{\ell}$ is a categorical distribution over classes $k\in\{1,2,\cdots, K\}$ associated with the leaf node $\ell$
and so $\sum_{k=1}^K(\pi_{\ell})_k=1$.
The code is in \href{https://github.com/tomsal/endtoenddecisiontrees}{End-to-end Learning of Deterministic Decision Trees}.

\subsection{Mondrian trees and forests}

Mondrian trees and forests are based on 
\href{https://www.researchgate.net/publication/221620618_The_Mondrian_Process}{The Mondrian Process},
which is a guillotine-partition-valued stochastic process.
The split functions of these models are generated on random.

In Chapters 5 and 6 of \cite{lakshminarayanan2016decision}, Mondrian forests are discussed,
where Bayesian inference is performed over leaf node parameters.

\href{https://papers.nips.cc/paper/3622-the-mondrian-process.pdf}{The Mondrian Process} can 
be interpreted as probability distributions over kd-tree data structures.
In another word, the samples drawn from Mondrian Processes are kd-trees. 
For example, a Mondrian process $m \sim MP(\lambda, (a,A), (b,B))$ is 
given a constant $\lambda$ and the rectangle $(a,A)\times (b,B)$.

\begin{algorithm}[h]
	\caption{Mondrian Processes}
	\begin{algorithmic}[1]
		\STATE {Input the specification  $\{\lambda, (a,A), (b,B)\}$}
		\STATE{Draw $E$ from an exponential distribution $Exp(A-a+B-b)$};
		\IF {$E\geq \lambda$}
		
		\STATE An axis-aligned cut is made uniformly at random along the combined lengths of $(a,A)$ and $(b,B)$:
		\STATE  Choose dimension $d$ with probability $\propto u_d-\ell_d$;
		\STATE  Choose cut location uniformly from $\ell_d, u_d$;
		\STATE  Recurse on left and right subtrees with parameter $\lambda- E$.
		\ELSE
		\STATE the process halts, and returns the trivial partition $(a,A)\times (b,B)$;    
		\ENDIF
	\end{algorithmic}
\end{algorithm}
More on the Mondrian process:
\href{https://papers.nips.cc/paper/3622-the-mondrian-process.pdf}{The Mondrian Process},
\href{https://arxiv.org/abs/2002.00797}{Stochastic geometry to generalize the Mondrian Process},
\href{https://cs.gmu.edu/~pwang7/rjmcmcMP.html}{Reversible Jump MCMC Sampler for Mondrian Processes}.

Similar to the extremely randomized trees, the learning of Mondrian trees consist of two steps:
\begin{enumerate}
	\item The split feature index $f$ is drawn with a probability proportional to $u_b[f] - l_b[f]$ where $u_b$ and $l_b$ and the upper and lower bounds of all the features.
	\item After fixing the feature index, the split threshold $\delta$ is then drawn from a uniform distribution with limits $l_b, u_b$.
\end{enumerate}
In another word, the learning of Mondrian trees are a sample drawn form the Mondrian process.

The prediction step of a Mondrian Tree is different from the classic decision trees.  
It takes into account all the nodes in the path of a new point from the root to the leaf for making a prediction. 
This formulation allows us the flexibility to weigh the nodes on the basis of how sure/unsure we are about the prediction in that particular node.

Mathematically, the distribution of $P(Y|X)$ is given by
$$P(Y\mid X)=\sum_j w_j \mathcal{N}(m_j,v_j)$$
where the summation is across all the nodes in the path from the root to the leaf. 
The mean prediction becomes $\sum_{j} w_j m_j$.

Formally, the weights are computed as following.
If $j$ is not a leaf, $w_j(x) = p_j(x) \prod_{k \in anc(j)} (1 - p_k(x))$, 
where  the first one being the probability of splitting away at that particular node 
and the second one being the probability of not splitting away till it reaches that node. 
If $j$ is a leaf, to make the weights sum up to one, $w_j(x) = 1 - \sum_{k \in anc(j)} w_k(x)$.
Here $p_j(x)$ denotes the probability of a new point $x$ splitting away from the node $j$; 
$anc(j)$ denotes the ancestor nodes of the node $j$.
We can observe that for $x$ that is completely within the bounds of a node, 
$w_j(x)$ becomes zero and for a point where it starts branching off, $w_j(x)=p_j(x)$.
So the Mondrian tree is expressed in the following form
\begin{equation}\label{Mondrian}
P(Y\mid X)=\sum_j p_j(x) \prod_{k \in anc(j)} (1 - p_k(x)) \mathcal{N}(m_j,v_j).
\end{equation}

The separation $p_j(x)$ of each node  is computed in the following way.
\begin{enumerate}
	\item $\Delta_{j} = \tau_{j} - \tau_{parent(j)}$;
	\item $\eta_{j}(x) = \sum_{f}(\max(x[f] - u_{bj}[f], 0) + \max(0, l_{bj}[f] - x[f]))$;
	\item $p_j(x) = 1 - e^{-\Delta_{j} \eta_{j}(x))}$.
\end{enumerate}

The implementation is in \href{https://scikit-garden.github.io/examples/MondrianTreeRegressor/}{Mondrian Tree Regressor in scikit-garden},
\href{https://github.com/balajiln/mondrianforest}{code on Modrian forests},
\href{https://github.com/harveydevereux/Mondrian}{Mondrian}, 
\href{https://indico.math.cnrs.fr/event/4132/attachments/2201/2697/scornet.pdf}{Scornet talk in Mondrian Tree},
\href{https://kernel.rubikloud.com/data-science/the-mondrian-process-digging-out-the-roots-of-mondrian-trees/}{a blog on mondrian-trees}.

Mondrian Forests is the sum of Mondrian trees such as 
\href{https://arxiv.org/abs/1406.2673}{Mondrian Forests},
\href{https://arxiv.org/abs/1906.10529}{AMF: Aggregated Mondrian Forests for Online Learning}.

An variant of Mondrian forest is the \href{http://papers.nips.cc/paper/9153-random-tessellation-forests.pdf}{Random Tessellation Forests}.

\section{Bayesian  decision trees}


As known, multiple adaptive regression spline is an extension of decision tree.  
And decision trees are a two-layer network. Both fields introduce the Bayesian methods.
It is necessary  to blend Bayesian methods and decision trees.
And it is important to introduce probabilistic methods to deal with the uncertainties in decision trees.
Recent history has seen a surge of interest in Bayesian techniques for constructing decision tree ensembles \cite{linero2017review}.

The basic idea of Bayesian thought of tree-based methods is to have the prior induce a posterior distribution 
which will guide a search towards more promising tree-based models 
such as \cite{chipman1998bayesian, denison1998bayesianc,  wu2007bayesian, nuti2019bayesian, hahn2020bayesian, chipman2010bart, hernandez2018bayesian, he2018xbart}.

\subsection{Bayesian trees}

From an abstract viewpoint, decision tree is a mapping $\mathbb{D}\to \mathbb{R}$ with the parameters $\Theta$ and $d$,
where $T$ guide the inputs to the terminal nodes and $\Theta$ is the parameters of prediction associated with the terminal nodes.
The Bayesian statisticians would pre-specify a prior on the parameters $\Theta$ to reflect the preferences or belief on different models.
Let $f(x\mid \Theta, T):x\mapsto Y$ be a decision tree.
In contrast to the conventional classification decision tree, the decision tree $f$ outputs the probability distribution of targets.
The conventional decision tree outputs the Dirac distribution or $\delta$ function on the mode of the targets associated with the terminal nodes.

In the ensemble perspectives, Bayesian tree is a posterior
$$\begin{cases}
\sum_{\Theta}f(x\mid \Theta, T) P(\Theta), &\text{if $\Theta$ is discrete;}\\
\int_{\Theta} T(x\mid \Theta, T) \mathrm{d}F(\Theta), &\text{if $\Theta$ is continuous.}
\end{cases}$$ 
Here $F(\Theta)$ is the cumulative probability function of the random variable $\Theta$.
As in \cite{nuti2019bayesian}, the output $Y$ is the probability drawn from the distribution $Y_x$
where the distribution of $Y_x$ will determine the type of problem we are solving: 
a discrete random variable translates into a classification problem 
whereas a continuous random variable translates into a regression problem.
In another word, the treed model then associates a parametric model for $Y$ at each of the terminal nodes of $T$.
More precisely, for values $x$ that are assigned to the $i$th terminal node of $T$, 
the conditional distribution of $Y\mid x$ is given by a parametric model
$$Y=f(x\mid \Theta, T)={P}(y\mid x, \Theta, T)\sim f(y\mid x, {\Theta}_i, T).$$
All points in the same set of a leaf node $\Pi$ share the same outcome distribution, 
i.e. $Y$ does not depend on $x$ but $\Theta, T$.
By using a richer structure at the terminal nodes, the treed models in a sense transfer structure from the tree to the terminal no des.

As presented in \cite{chipman1998bayesian}, association of the individual $Y$ values with the terminal nodes is indicated 
by letting $Y_{ij}$ denote the $j$th observation
of $Y$ in the $i$th partition (corresponding to the $i$th terminal node), $i \in \{1,2,\cdots, b\}, j\in\{1,2,\cdots, n_i\}.$
In this case, the CART model distribution for the data will be of the form
$$P(Y\mid X,\Theta)=\prod_{i=1}^{b}f(Y_i\mid \theta_i)=\prod_{i=1}^{b}\prod_{j=1}^{n_i}f(Y_{ij}\mid \theta_i)$$
where  we use $f$ to represent a parametric family indexed by $\theta_i$.
Since a CART model is identified by $(\Theta, T)$, a Bayesian analysis of the problem proceeds by specifying a prior probability distribution $P(\Theta, T)$.
This is most easily accomplished by specifying a prior $P(T)$ on the tree space and a conditional prior $P(\theta\mid T)$ on the parameter space, 
and then combining them via $P(\Theta, T)=P(T)P(\Theta\mid T)$.
The features are also pointed out in\cite{chipman1998bayesian}:
(1) the choice of prior for $T$ does not depend on the form of the parametric family indexed by $\Theta$;
(2) the conditional specification of the prior on $\Theta$ more easily allows for the choice of convenient analytical forms 
which facilitate posterior computation.

In \cite{chipman1998bayesian}, the tree prior $P(T)$ is implicitly specified by a tree-generating stochastic process.
As in \cite{chipman2002bayesian}, this prior is implicitly defined by a tree-generating stochastic process 
that grows trees from a single root tree by randomly splitting terminal nodes. 
A tree's propensity to grow under this process is controlled by a two-parameter node splitting probability 
$$P(\text{node splits j}\mid depth = d) = \alpha(1 + d)^{-\beta},$$
where the root node has depth 0. 
The parameter $\alpha$ is a base probability of growing a tree by splitting a current terminal node 
and $\beta$ determines the rate at which the propensity to split diminishes as the tree gets larger.
The tree prior $P(T)$ is completed by specifying a prior on the splitting rules assigned to intermediate nodes.
The stochastic process for drawing a tree from this prior can be described in the following recursive manner:
\begin{enumerate}
  \item Begin by setting $T$ to be the trivial tree consisting of a single root (and terminal) node denoted $\eta$.
  \item Split the terminal node $\eta$ with probability $P(\text{node splits j}\mid depth = d)$.
  \item If the node splits, assign it a splitting rule $\rho$ according to the distribution $P_{RULE}(\rho\mid \eta, d)$, and create the left and right children nodes. 
  Let $T$ denote the newly created tree, and apply steps 2 and 3 with $\eta$ equal to the new left and the right children (if nontrivial splitting rules are available).
\end{enumerate}
Here $P_{RULE}(\rho\mid \eta, d)$ is a distribution on the set of available predictors $x_i$, and conditional on each predictor, 
a distribution on the available set of split values or category subsets.
As a practical matter, we only consider prior specification for 
which the overall set of possible split values is finite. 
Thus, each $P_{RULE}(\rho\mid \eta, d)$ will always be a discrete distribution.
Note also that because the assignment of splitting rules will typically depend on $X$, the prior $P(T)$ will also depend on $X$.

The specification of  $P(\Theta\mid T)$ will necessarily be tailored to the particular form of the model $P(y \mid x, \theta )$ under consideration.
In particular, a key consideration is to avoid conflict between $P(\Theta\mid T)$ and the likelihood information from the data.
In \cite{chipman1998bayesian, chipman2000hierarchical, chipman2002bayesian}, diverse priors were proposed.
Starting with an initial tree $T^0$, iteratively simulate the transitions from $T^i$ to $T^{i+1}$ by the two steps:
\begin{enumerate}
  \item Generate a candidate value $T^{\ast}$ with probability distribution $q(T^i,  T^{\ast})$.
  \item Set $T^{i+1}=T^{\ast}$ with probability
         $$\alpha(T^{i}, T^{\ast})=\min\{\frac{q(T^{\ast}, T^{i})p(Y\mid X, T^{\ast})p(T^{\ast})}{q(T^{i},T^{\ast})p(Y\mid X, T^i)p(T^i)}, 1\}.$$
   Otherwise, set $T^{i+1}=T^{i}$.
\end{enumerate}

We consider kernels $q(T^{\ast}, T^{i})$ which generate $T^{\ast}$ from $T$ by randomly choosing among four steps:
\begin{itemize}
  \item GROW: Randomly pick a terminal node. Split it into two new ones by randomly assigning it a splitting rule according to $p_{RULE}$ used in the prior.
  \item PRUNE: Randomly pick a parent of two terminal nodes and turn it into a terminal node by collapsing the nodes below it.
  \item CHANGE: Randomly pick an internal node, and randomly reassign it a splitting rule according to $p_{RULE}$ used in the prior.
  \item SWAP: Randomly pick a parent-child pair which are both internal nodes. Swap their splitting rules unless the other child has the identical rule. 
    In that case, swap the splitting rule of the parent with that of both children.
\end{itemize}

\subsection{Bayesian additive regression trees}

BART\cite{chipman2010bart, hernandez2018bayesian, he2018xbart} is a nonparametric Bayesian regression approach which
uses dimensionally adaptive random basis elements. 
Motivated by ensemble methods in general, and boosting algorithms in particular, BART is defined by a statistical model: a prior and a likelihood.
BART differs in both how it weakens the individual trees by instead using a prior, 
and how it performs the iterative fitting by instead using Bayesian back-fitting on a mixed number of trees.

We complete the BART model specification by imposing a prior over all the parameters of \eqref{Additive DT} including the total number of decision trees, all the bottom node parameters
as well as the tree structures and decision rules.
For simplicity, the tree components  are set to be independent of each other and of error, and the terminal node parameters of every tree are independent.
In \cite{chipman2010bart}, each tree follows the same prior as in \cite{chipman1998bayesian}.
And the errors are in normal distribution, i.e., $\epsilon \sim \mathcal{N}(0, \sigma)$ and $\sigma$ use conjugate prior the inverse chi-square distribution.

BART can also be used for variable selection by simply selecting those variables that appear most often in the fitted sum-of-trees models.

A modified version of BART is developed in \cite{he2018xbart}, which is amenable to fast posterior estimation.

The implementation of BART is in the project \href{http://www.github.com/theodds/SoftBART}{SoftBart}.
And we can use BART to casual inference \href{https://rdrr.io/cran/bartCause/man/bartc.html}{BARTC}.
\subsection{Bayesian MARS}

The Bayesian approach is applied to univariate and multivariate adaptive regression spline (MARS) such as \cite{denison1998bayesianm, holmes2003classification, francom2018sensitivity}.


Multivariate adaptive regression spline (MARS) \cite{fridedman1991multivariate}, motivated by the recursive partitioning approach to regression, is given by
\begin{equation}\label{MARS} \sum_{i=1}^{k}a_i B_i(\mathrm{x})\end{equation}
where $\mathrm{x} \in D$ and the $a_i(i = 1, \cdots, k)$ are the suitably chosen
coefficients of the basis functions $B_i$ and $k$ is the number of basis functions in the model.
The $B_i$ are given by 
\begin{equation}\label{BasisMARS}
B_i = \begin{cases}1, &\text{$i=1$}\\
    \prod_{j}^{J_i}[s_{ij}\cdot(x_{\mu(ij)}-t_{ij})]_{+}, &\text{$i=2,3,\cdots$}\end{cases}
\end{equation}
where $[x]_{+}=\max(0, x)$; $J_i$ is the degree of the interaction of basis $B_i$, 
the $s_{ji}$, which we shall call the sign indicators, equal $\pm 1$, 
the $\mu(ij)$ give the index of the predictor variable 
which is being split on the $t_{ji}$ (known as knot points) give the position of the splits. 
The $\mu(j, \cdot)(j=1, J)$ are constrained to be distinct so each predictor only appears once in each interaction term.
See Section 3, \cite{fridedman1991multivariate} for more details.

\begin{table}   
  \caption{Comparison of decision trees and multivariate adaptive regression spline}  
  \begin{center}  
  \begin{tabular}{p{4cm}|p{4cm}|p{4cm}}
  \hline  
  Methods & Decision Trees & MARS \\
  \hline  
  \hline
  Non-terminal functions & Test functions         & Truncation functions \\
  \hline
  Terminal functions     & Constant  functions    & Polynomial function \\
  \hline
  Training methods       & Induction and Pruning  &  Forwards search and backwards deletion\\
  \hline
  Trained results        & Simple functions       & Piece-wise polynomial functions\\
  \hline
  Routing                & Determined routing     & No routing \\
  \hline
  Hyperparameter         & the number of terminal nodes & the number of basis functions\\
  \hline
  \end{tabular}  
  \end{center}
\end{table}

Bayesian MART\cite{denison1998bayesianm} set up a probability distribution over the space of possible MARS structures.
Any MARS model can be uniquely defined by the number of basis functions present, the coefficients and the types of the basis functions, together with the knot
points and the sign indicators associated with each interaction term.
In another word, it is $k, a_i, s_{ij}, t_{ij}$ \eqref{BasisMARS}, \eqref{MARS} and the types of basis function $B_i$
Here the type of basis function $B_i$ is described by $T_i$, 
which just tells us which predictor variables we are splitting on, 
i.e. what the values of $\mu(1, i),\cdots, \mu(J_i, i)$ are.

A truncated Poisson distribution (with parameter $k$) is used to specify the prior probabilities for the number of basis functions, giving
$$p(k\mid \lambda)=\frac{\lambda^k}{\alpha k!}, \forall k\in\{1,2,\cdots, k_{max}\}$$
where $\alpha$ is the normalization constant.

We can apply this Bayesian scheme directly into decision trees based on the representation \eqref{sum-product} in \cite{fridedman1991multivariate}.


\section{Optimal decision trees}

Generally, the decision tree learning methods are greedy and recursively as shown in \cite{holzinger2015data, loh2014fifty}.
Here we will focus on numerical optimization methods for decision trees.
The optimal decision trees are expected as the best multivariate tree with a given
structure.
However, such optimum condition is difficult to verify because of their comibinatorical structure.
So we need optimization techniques to minimize the error of the entire decision tree instaed of the greedy or heuristic induction methods.
\subsection{Optimal classification trees}

Bertsimas Dimitris and Dunn Jack \cite{bertsimas2017optimal}
present Optimal Classification Trees, a novel formulation of the decision tree problem using
modern MIO techniques that yields the optimal decision tree.
The core idea of the MIP formulation of optimal decision tree is  
\begin{itemize}
  \item to construct the maximal tree of the given depth;
  \item to enforce the requirement of decision trees with the constraints;
  \item to minimize the cost function subject to the constraints.
\end{itemize}

\begin{table}  
  \caption{The Notation of MIP Formulation of Decision Trees}  
  \begin{center}  
  \begin{tabular}{p{3cm}|p{7cm}}
  \hline  
  Notation  & Definition \\
  \hline
  $N_{min}$ &  the minimum number of points of all nodes\\
  $N_{x}(l)$ & the number of training points contained in leaf node $l$\\ 
  $(x_i, y_i)$ & the training data, where $x_i\in[0, 1]^p, y_i\in\{1,\cdots, K\}$ for $i = 1, \cdots, n.$ \\
  $p(t)$ &  the parent node of node $t$\\
  $A(t)$ &  the set of ancestors of node $t$\\
  $A_L(t)$ & the set of ancestors of $t$ whose left branch has been followed on the path from the root node to $t$\\
  $A_R(t)$ & the set of right-branch ancestors of $t$, $A(t)=A_L(t)\cup A_R(t)$\\
  $\mathcal{T}_{B}$  & branch node set, also a.k.a. non-terminal nodes, apply a linear split\\
  $\mathcal{T}_{L}$  & leaf node set, also a.k.a. terminal nodes, make a class prediction  \\
  $a_t\in\mathbb{R}^p$ & the combination coefficients of the split function at the node $t$ \\
  $b_t\in\mathbb{R}$ & the threshold of the split function at the node $t$ \\
  $d_t$& $d_t=\mathbb{I}(\text{the node t  applies a split})$, the indicator variables  to track which branch nodes apply splits\\
  $z_{it}$ & $z_{it}=\mathbb{I}(x_i \text{ in the node t })$,the indicator variables  to track the points assigned to each leaf node $t$\\
  $l_{t}$ & $l_{t}=\mathbb{I}(\text{the node t is not null})$,the indicator variables  to track cardinality of each leaf node \\
  $M_1, M_2$ & big constants\\
  $N_{kt}$ & the number of points of label $k$ in node $t$, \\
  $N_t$ & be the total number of points in node $t$\\
  $\{ 1, \cdots , K\}$ & the ground truth set\\
  $N_t$ & the total number of points in node $t$ \\
  $Y_{ik}$ &  Kronecker delta function \\
  $N_{kt}$ &  the number of points of label $k$ in node $t$ \\
  $c_{kt}$ & the variable to track the prediction of the node $t$\\
  \hline
  \end{tabular}  
  \end{center}  
\end{table}
It is first to discuss the axes-aligned/univariate decision tree.
If the depth $D$ is given, we will determine the structure of the maximal tree for this depth, 
i.e., the variables $A(t), A_L(t), A_R(t)$ and  $\mathcal{T}_{B}, \mathcal{T}_{L}$.
If a branch node does apply a split, i.e. $d_t=1$, then the axes-aligned split $a_t$ will select one and only one attribute to test;
If a branch node does not apply a split, i.e. $d_t=0$, then the axes-aligned split $a_t$ is set to be $\vec{0}$ 
so that all points are enforced to follow the right split at this node.\footnotemark
\footnotetext{Here the left split of the node $t$ is $a_tx<b_t$ and the right split is $a_tx\leq b_t$.}
These are enforced with the following constraints on split functions:
$$\sum_{j=1}^p a_{jt} =d_t,\forall t\in\mathcal{T}_{B},$$
$$0\leq b_t\leq d_t\leq 1, \forall t\in\mathcal{T}_{B},$$
$$a_{jt}\in\{0, 1\},\forall t\in\mathcal{T}_{B}.$$

The hierarchical structure of the tree are enforced via 
$$d_t\leq d_{p(t)}\quad\forall t\in \mathcal{T}_{B}-\{ 1\}$$
which means that the node $t$ is likely to apply a split function only if its parent applies a split function.
The above constraints are not related with the information of samples and 
they are just necessary requirement on the decision trees.

We also force each point to be assigned to exactly one leaf
$$\sum_{t\in \mathcal{T}_{L}}z_{it}=1, i=1,2,\cdots, n$$
and each leaf has at least the minimum number of samples
$$z_{it}\leq l(t),$$
$$\sum_{i=1}^{n}z_{it}\geq N_{min} l_{t},t\in\mathcal{T}_{L}\cup \mathcal{T}_{L}.$$
Finally, we apply constraints enforcing the splits that are required by the structure of the
tree when assigning points to leaves
$$a_m^Tx_i+\epsilon\leq b_m + M_1 (1 - z_{it}), i = 1, \cdots, n, \forall t\in\mathcal{T}_{B}, \forall m\in A_L(t)$$
$$a_m^Tx_i\geq b_m - M_2 (1 - z_{it}), i = 1, \cdots, n, \forall t\in\mathcal{T}_{B}, \forall m\in A_R(t)$$
The largest valid value of $\epsilon$ is the smallest non-zero distance between adjacent values of this feature, i.e.,
$$\epsilon+i=\max\{x_j^{i+1}-x_j^{(i)}\mid x_j^{(i+1)}\neq x_j^{(i)},i=1,\cdots, n,\}$$
where $x_j^\{(i)\}$ is the $i$-th largest value in the $j$-th feature.

The total number of points in node $t$ and the number of points of label $k$ in node $t$:
$N_{kt}=\frac{1}{2}\sum_{i=1}^{n} (Y_{ik}+1)z_{it}, k= 1, \cdots , K, t\in \mathcal{T}_{L},$ and 
$$N_t=\sum_{k=1}^{K} N_{kt}=\sum_{i=1}^n z_{it}, t\in \mathcal{T}_{L}.$$

The optimal label of each leaf to predict is:
$$c_t=\arg\max_{k\in\{1,2,\cdots, K\}}N_{kt}$$
and we use binary variables $c_{kt}$ to track the prediction of each node where  
$$c_{kt}=\mathbb{I}_{c_t =k}=\begin{cases} 1, &\text{if $c_t=k$}\\
0,&\text{otherwise}\end{cases}.$$
We must make a single class prediction at each leaf node that contains points:
$$\sum_{k=1}^{K}c_{kt}=l_t, \forall t\in \mathcal{T}_{L}.$$
The misclassification cost is 
$$L_t=N_t-N_{c_t\,\, t}=N_t-\max_{k\in\{1,2,\cdots, K\}}N_{kt}=\min_{k\in\{1,2,\cdots, K\}}N_t-N_{kt}$$
which can be linearized to give
$$L_t\geq N_t - N_{kt} - M(1 - c_{kt} ), k = 1, \cdots , K,\forall t\in \mathcal{T}_{L}$$
$$L_t\leq N_t - N_{kt} - M c_{kt}, k = 1, \cdots , K,\forall t\in \mathcal{T}_{L}$$
$$L_t\geq 0,\forall t\in \mathcal{T}_{L}.$$
where again $M$ is a sufficiently large constant that makes the constraint inactive depending on the value of $c_{kt}$

$$
\begin{aligned}
 & \min \sum_{t\in \mathcal{T}_{L}}L_t + \alpha \sum_{t\in \mathcal{T}_{B}} d_t\\
 & \text{subject to  } L_t\geq N_t - N_{kt} - M(1 - c_{kt} ), k = 1, \cdots , K,\forall t\in \mathcal{T}_{L}\\
 & L_t\leq N_t - N_{kt} - M c_{kt}, k = 1, \cdots , K,\forall t\in \mathcal{T}_{L},
  L_t\geq 0,\forall t\in \mathcal{T}_{L} \\
 & N_{kt}=\frac{1}{2}\sum_{i=1}^{n} (Y_{ik}+1)z_{it}, k= 1, \cdots , K, t\in \mathcal{T}_{L}, \\
 & N_t=\sum_{k=1}^{K} N_{kt}=\sum_{i=1}^n z_{it}, t\in \mathcal{T}_{L},
 \sum_{k=1}^{K}c_{kt}=l_t, \forall t\in \mathcal{T}_{L},\\
 & a_m^Tx_i+\epsilon\leq b_m + M_1 (1 - z_{it}), i = 1, \cdots, n, \forall t\in\mathcal{T}_{B}, \forall m\in A_L(t),\\
 & a_m^Tx_i\geq b_m - M_2 (1 - z_{it}), i = 1, \cdots, n, \forall t\in\mathcal{T}_{B}, \forall m\in A_R(t),\\
 & \sum_{j=1}^p a_{jt}=d_t,\forall t\in\mathcal{T}_{B},
0\leq b_t\leq d_t\leq 1, \forall t\in\mathcal{T}_{B},\\
 & a_{jt}\in\{0, 1\},\forall t\in\mathcal{T}_{B}, d_t\leq d_{p(t)}\forall t\in \mathcal{T}_{B}-\{ 1\},\\
 & \sum_{t\in \mathcal{T}_{L}}z_{it}=1, i=1,2,\cdots, n,
  z_{it}\leq l(t),\sum_{i=1}^{n}z_{it}\geq N_{min} l_{t},t\in\mathcal{T}_{L}\cup \mathcal{T}_{L}.
\end{aligned}
$$

Note that the indicator variables such as $d_{i}, z_{it}$ are in $\{0, 1\}$ and $b_m\in\mathbb{R}$
so it is a mixed integer optimization problem.

For more on optimal decision trees see \href{https://ideas.repec.org/a/inm/oropre/v55y2007i2p252-271.html}{Dimitris Bertsimas and Romy Shioda}\cite{bertsimas2007classification},
\href{http://genoweb.toulouse.inra.fr/~tschiex/CP2019/paper132post.pdf}{H\`{e}l\`{e}ne Verhaeghe et al}\cite{DBLP:conf/bnaic/VerhaegheNPQS19},
\href{http://www.optimization-online.org/DB_FILE/2018/01/6404.pdf}{Oktay et al}\cite{DBLP:journals/corr/MenickellyGKS16},
\href{https://www.researchgate.net/publication/2796065_Optimal_Decision_Trees}{Kristin P. Bennettand and Jennifer A. Blue}\cite{bennett1996optimal},
\href{http://papers.nips.cc/paper/7716-boolean-decision-rules-via-column-generation.pdf}{Sanjeeb Dash et al}\cite{DBLP:conf/icassp/DashMV15},
\href{https://link.springer.com/chapter/10.1007/978-3-319-59776-8_8}{Sicco Verwer and Yingqian Zhang}\cite{verwer2017learning, DBLP:conf/aaai/VerwerZ19},
\href{https://arxiv.org/pdf/1810.06684.pdf}{Murat Firat et al}\cite{DBLP:journals/corr/abs-1810-06684}.

\subsection{Tree alternating optimization}

\href{https://papers.nips.cc/paper/7397-alternating-optimization-of-decision-trees-with-application-to-learning-sparse-oblique-trees.pdf}{TAO}
\cite{carreira2018alternating} cycle over depth levels from the bottom (leaves) to the top (root) and iterate bottom-top, bottom-top, etc. 
(i.e., reverse breadth-first order).
TAO can actually modify the tree structure by ignoring the  dead branches and pure subtrees.
And it shares the indirect pruning with MIO.
\begin{table}  
  \caption{Comparison of MIO and TAO for Decision Trees}  
  \begin{center}  
  \begin{tabular}{p{5cm}|p{5cm}}
  \hline  
   MIO & TAO \\
  \hline  
  \hline
  maximum tree of given depth          & given tree structure \\
  \hline
  $a_i=\vec{0}, b_i=0$                 & Dead branches \\
  \hline
  decision function                    & split function \\
  \hline
  all their points have the same label & Pure subtrees \\
  \hline
  $d_t\leq d_{p(t)}\forall t\in \mathcal{T}_{B}-\{ 1\}$ & dead branches and pure subtrees are ignored\\
  \hline
  by optimizing the coefficients of the split functions of the nodes & by reducing the size of the tree to modify the tree structure\\
  \hline
  specified blood relatives of the nodes & separability condition \\
  \hline
  \end{tabular}  
  \end{center}
\end{table}

We want to optimize the the parameters of all nodes in the tree to minimize the misclassification cost jointly
$$L(T(x\mid\Theta))=\sum_{i=1}^n\ell(T(x_i), y_i)$$
where $\Theta=\{a_t\in\mathbb{R}^p,b_t\in\mathbb{R}, c_l\mid \forall i\in \mathcal{T}_{B}, l\in\mathcal{T}_{L}\}$.
The separability condition allows us to optimize separately (and in parallel) over the parameters
of any set of nodes that are not descendants of each other (fixing the parameters of the remaining
nodes).
Optimizing the misclassification cost over an internal node $t$ is exactly equivalent to a reduced problem:
a binary misclassification loss for a certain subset $C_t$ (defined below) of the training points over the parameters of the test function.

Firstly, optimizing the misclassification error over $\theta_t=(a_t,b_t)$ in  the misclassification cost, 
where is summed over the whole training set, 
is equivalent to optimizing it over the subset of training points $S_t=\{x_i\mid z_{it}=1, i=1,\cdots,n\}$ that reach node $t$.

Next, the fate of a point $x\in S_t$ depends only on which of $t$’s children it follows
because other parameters are fixed.
The modification of the  decision function will only change some paths of samples in $S_t$ called as altered travelers,
which only change partial results of the prediction.
In another word, some altered travelers eventually reach the different leaf nodes with previous prediction
while some altered travelers eventually reach the same leaf as previously.

Then, we can define a new, binary classification problem over the parameters $\theta_t$ of the decision function on the altered travelers.
Thus it is to optimize a reduced problem.

There is no approximation guarantees for TAO at present. 
TAO does converge to a local optimum in the sense of alternating optimization (as in $k$-means), i.e., when no
more progress can be made by optimizing one subset of nodes given the rest.

For more information on TAO, see the following links.
\begin{enumerate}
  \item \href{https://faculty.ucmerced.edu/mcarreira-perpinan/research/TAO.html}{Mcarreira Perpinan's research on TAO};
  \item \href{https://faculty.ucmerced.edu/mcarreira-perpinan/papers/neurips18-supp/neurips18-supp.pdf}{Supplementary material for TAO};
  \item \href{https://arxiv.org/abs/1911.03054}{An Experimental Comparison of Old and New Decision Tree Algorithms}.
\end{enumerate}

\subsubsection{Surrogate objective}

\href{https://arxiv.org/pdf/1506.06155.pdf}{Mohammad Norouzi, Maxwell D. Collins, David J. Fleet, Pushmeet Kohli} 
propose a novel algorithm for optimizing multivariate 
linear threshold functions as split functions of decision trees to create improved Random Forest classifiers.
\href{https://norouzi.github.io/}{Mohammad Norouzi}, \href{https://scholar.google.com/citations?user=V0hkNigAAAAJ&hl=en}{Maxwell D. Collins},
\href{http://www.cs.toronto.edu/~fleet/}{David J Fleet} et al  introduce a tree \href{https://arxiv.org/pdf/1511.04056.pdf}{navigation function}
$f:\mathbb{H}^m \mapsto \mathbb{I}^{m+1}$ that maps an $m$-bit sequence of split decisions ($\mathbb{H}^m = \{-1, +1\}^m$) to an indicator
vector that specifies a 1-of-$(m + 1)$ encoding.
However, it is not to formulate the dependence of navigation function on binary tests.
It is shown that the problem of finding optimal linear-combination (oblique) splits for decision trees 
is related to structured prediction with latent variables.
The  empirical loss function is  re-expressed as
\begin{equation}\label{loss}
 L(S, v;\mathcal{D})=\sum_{(x, y)\in\mathcal{D}}\ell(v^Tf(\hat{h}(x)), y),\\
s.t. \hat{h}(x)=\arg\max_{h\in\mathbb{H}^m}(h^TSx)
\end{equation}
where $f$ is the navigation function.
And $\hat{h}(x)_i=\frac{(Sx)_i}{|(Sx)_i|}$.
And an upper bound on loss for an input-output pair, $(x; y)$ takes the form
\begin{equation}\label{Upper bound}
\ell(v^T f(\underbrace{\operatorname{sgn}(Sx))}_{\max_{h\in\mathbb{H}^m}(h^TSx)}, y)\leq 
\max_{g\in\mathbb{H}^m}(g^T Sx+\ell(v^T f(g)), y)-\max_{h\in\mathbb{H}^m}(h^TSx).
\end{equation}
To develop a faster loss-augmented inference algorithm, we formulate
a slightly different upper bound on the loss, i.e.,
\begin{equation}\label{ Augmented upper bound}
\ell(v^T f(\operatorname{sgn}(Sx)), y)\leq 
\max_{g\in\in\mathcal{B}_1(\operatorname{sgn}(Sx)}(g^T Sx+\ell(v^T f(g)), y)-\max_{h\in\mathbb{H}^m}(h^TSx).
\end{equation}
where $\mathcal{B}_1(\operatorname{sgn}(Sx)$ denotes the Hamming ball of radius 1 around $\operatorname{sgn}(Sx)$.
And a regularizer is introduced  on the norm of $W$ when optimizing the bound
\begin{equation} 
\begin{split}
&\max_{g\in\mathbb{H}^m}(ag^T Sx+\ell(v^T f(g)), y)-\max_{h\in\mathbb{H}^m}(ah^TSx)\leq \\
&\max_{g\in\mathbb{H}^m}(bg^T Sx+\ell(v^T f(g)), y)-\max_{\in\mathbb{H}^m}(bh^TSx).
\end{split}
\end{equation}
for $a> b> 0$.
Summing over the bounds for different training pairs and constraining the norm of rows of $S$, we the surrogate objective:
\begin{equation}\label{surrogatr loss}
\begin{split}
& L^{\prime}(S, v;\mathcal{D})=\sum_{(x, y)\in\mathcal{D}}(\max_{g\in\mathbb{H}^m}(bg^T Sx+\ell(v^T f(g)), y)-\max_{h\in\mathbb{H}^m}(bh^TSx))\\
&\text{s.t.  } \|s_i\|_2\leq \nu \quad\forall i=\{1,2,\cdots, m.\}
\end{split}  \end{equation}
where $\nu\in\mathbb{R}^{+}$ is a regularization parameter and $s_i$ is the $i$ row of $S$.

We observe that after several gradient updates some of the leaves may end up not being assigned to any data points and 
hence the full tree capacity may not be exploited. 
We call such leaves inactive as opposed to active leaves that are assigned to at least one training data point. 
An inactive leaf may become active again, but this rarely happens given the form of gradient updates. 
To discourage abrupt changes in the number of inactive leaves, we introduce a variant of SGD, 
in which the assignments of data points to leaves are fixed for a number of gradient update steps. 
Thus, the bound is optimized with respect to a set of data point to leaf assignment constraints. 
When the improvement in the bound becomes negligible the leaf assignment variables are updated, 
followed by another round of optimization of the bound. 
We call this algorithm Stable SGD (SSGD) 
because it changes the assignment of data points to leaves more conservatively than SGD.

Based on this surrogate objective, a Maximization-Minimization type method is proposed as following
\begin{equation}\begin{split}
g &=\arg\max_{g\in\mathbb{H}^m}\{(g^T Sx+\ell(v^T f(g)), y)-\max_{h\in\mathbb{H}^m}(h^TSx)\}\\
v &=\arg\min_{v\in\mathbb{R}^m}\{\max_{g\in\mathbb{H}^m}(g^T Sx+\ell(v^T f(g)), y)-\max_{h\in\mathbb{H}^m}(h^TSx)\}
\end{split}
\end{equation}

\subsection{Differentiable trees}

As put in \href{https://pure.tudelft.nl/portal/files/66632115/Hehn2019_Article_End_to_EndLearningOfDecisionTr.pdf}{End-to-End Learning of Decision Trees and Forests}
\begin{quote}
One can observe that both neural networks and decision trees are composed of basic computational units, the perceptrons and nodes, respectively. 
A crucial difference between the two is that in a standard neural network, all units are evaluated for every input, 
while in a reasonably balanced decision tree with $I$ inner split nodes, only $O(\log I)$ split nodes are visited. 
That is, in a decision tree, a sample is routed along a single path from the root to a leaf, with the path conditioned on the sample’s features.
\end{quote}

Here  we refer differentiable trees to the soft decision trees which can be trained  via gradient-based methods.

\href{https://arxiv.org/pdf/1712.02743.pdf}{End-to-end Learning of Deterministic Decision Trees},
\href{https://pure.tudelft.nl/portal/files/66632115/Hehn2019_Article_End_to_EndLearningOfDecisionTr.pdf}{End-to-End Learning of Decision Trees and Forests},
\href{https://www.microsoft.com/en-us/research/wp-content/uploads/2016/06/ICCV15_DeepNDF_main.pdf}{Deep Neural Decision Forests}
``soften” decision functions in the internal tree nodes to make the overall tree function
and tree routing differentiable.

\begin{table}  
  \caption{Comparison of decision trees and deep neural networks}  
  \begin{center}  
  \begin{tabular}{p{5cm}|p{5cm}}
  \hline  
   Decision Trees & Deep Neural Networks \\
  \hline  
  \hline
  Test functions & Nonlinear activation functions \\
  \hline
  Node to node & Layer by layer \\
  \hline
  Recursively-partitioning-based growth methods &  Gradient-based optimization methods\\
  \hline
  Logical transparency & High numerical efficiency\\
  \hline
  Probabilistic decision trees & Bayesian Deep Learning \\
  \hline
  Decision stump & Perception \\
  \hline
  Decision tree & A hidden-layer neural network \\
  \hline
  \end{tabular}  
  \end{center}
\end{table}

\subsubsection{Deep neural decision trees}
Yongxin Yang et al\cite{yang2018deep} construct the decision tree via Kronecker product $\otimes$
$$z=f_1(x_1)\otimes f_2(x_2)\otimes\cdots\otimes f_D(x_D)$$
where each feature $x_d$ is binned by its own neural network $f_d(x_d)$.
Here $f_d(\cdot)$ is a one-layer neural network with softmax as its activation function:
$$\operatorname{softmax}(\frac{wx+b}{\tau})$$
where $x\in\mathbb{R}$, $w = [1,2,\cdots, n+1]$, $b=[0, -\beta_1, -\beta_1-\beta_2,\cdots, \sum_{i=1}^{n}-\beta_i]$
and $\tau>0$ is a temperature factor.
As $\tau\to\infty$, the output tends to a one-hot vector.
Here $z$ is now also an almost one-hot vector 
that indicates the index of the leaf node where instance $x$ arrives. 
Finally, we assume a linear classifier at each leaf $z$ classifies instances arriving there.

In summary, it is to bin the features instead of the recursive partitioning as usual.
For more details of its implementation see \UrlBigBreaks{https://github.com/wOOL/DNDT}.

\subsubsection{Deep neural decision forests}

\href{https://www.microsoft.com/en-us/research/wp-content/uploads/2016/06/ICCV15_DeepNDF_main.pdf}{Deep Neural Decision Forests}
unifies classification trees with the representation learning functionality known from deep convolutional networks.
Each decision node (branching node as in MIO) is responsible for routing samples along the tree.
When a sample  reaches a decision node $n$ it will be sent to the left or right subtree based on the output of decision/test/split function. 
In standard decision forests, the decision function is binary and the routing is deterministic.
In order to provide an explicit form for the routing function
we introduce the following binary relations that depend on the tree’s structure: 
\begin{itemize}
  \item $\ell \swarrow n$: if the leaf $\ell$ belongs to the left subtree of node $n$;
  \item $\ell\searrow n$: if the leaf $\ell$ belongs to the right subtree of node $n$.
\end{itemize}
We can now exploit these relations to express routing function $\nu_{\ell}(x\mid \Theta)$ providing the probability that sample $x$ will reach leaf $\ell$ as follows:
$$\mu_{\ell}(x\mid \Theta)=\prod_{n\in\mathcal{T}_B}d_{n}(x\mid \Theta)^{\mathbb{I}_{\ell \swarrow n}}\bar{d}_{n}(x\mid \Theta)^{\mathbb{I}_{\ell \searrow n}}$$
where $\bar{d}_{n}(x\mid \Theta)=1-d_{n}(x\mid \Theta)$ and $\mathbb{I}_{P}$ is an indicator function conditioned on the argument $P$.
And the final prediction for sample $x$ from tree $T$ with decision nodes parametrized by $\Theta$ is given by
$$\mathbb{P}_{T}[y\mid x,\Theta, \pi]=\sum_{\ell\in\mathcal{T}_L}\pi_{\ell y} \mu_{\ell}(x\mid \Theta)$$
where $\pi_{\ell y}$ denotes the probability of a sample reaching leaf $\ell$ to take on class $y$.
\begin{figure}[H] 
\centering 
\includegraphics[width=0.5\textwidth]{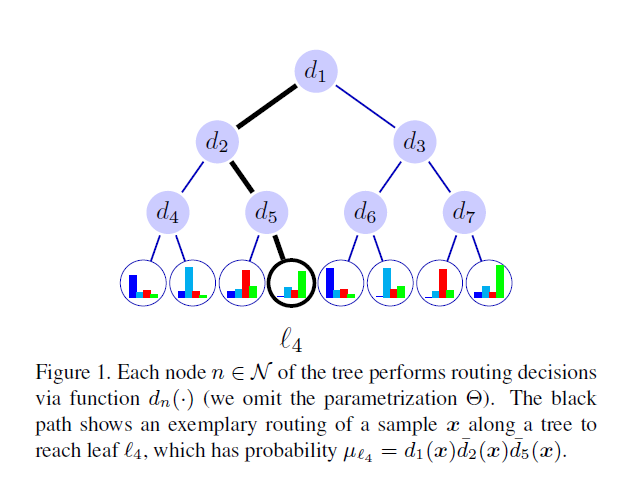}
\caption{The Stochastic Routing Diagram}  
\label{Fig.Routing} 
\end{figure}
The decision functions delivering a stochastic routing are defined as follows:
$$d_n(x\mid \Theta)=\sigma(f_n(x\mid \Theta))$$
where $\sigma(x)=\frac{1}{1+\exp(-x)}$ is the sigmoid function;
$f_n(x\mid \Theta):\mathcal{X}\mapsto \mathbb{R}$ is a real-valued function depending on the sample and the parametrization $\Theta$.
Our intention is to endow the trees with feature learning capabilities 
by embedding functions $f_n$ within a deep convolutional neural network with parameters $\Theta$.
In the specific, we can regard each function $f_n$ as a linear output unit of a deep network 
that will be turned into a probabilistic routing decision by the action of $d_n$, 
which applies a sigmoid activation to obtain a response in the $[0; 1]$ range.
\begin{figure}[H] 
\centering 
\includegraphics[width=0.85\textwidth]{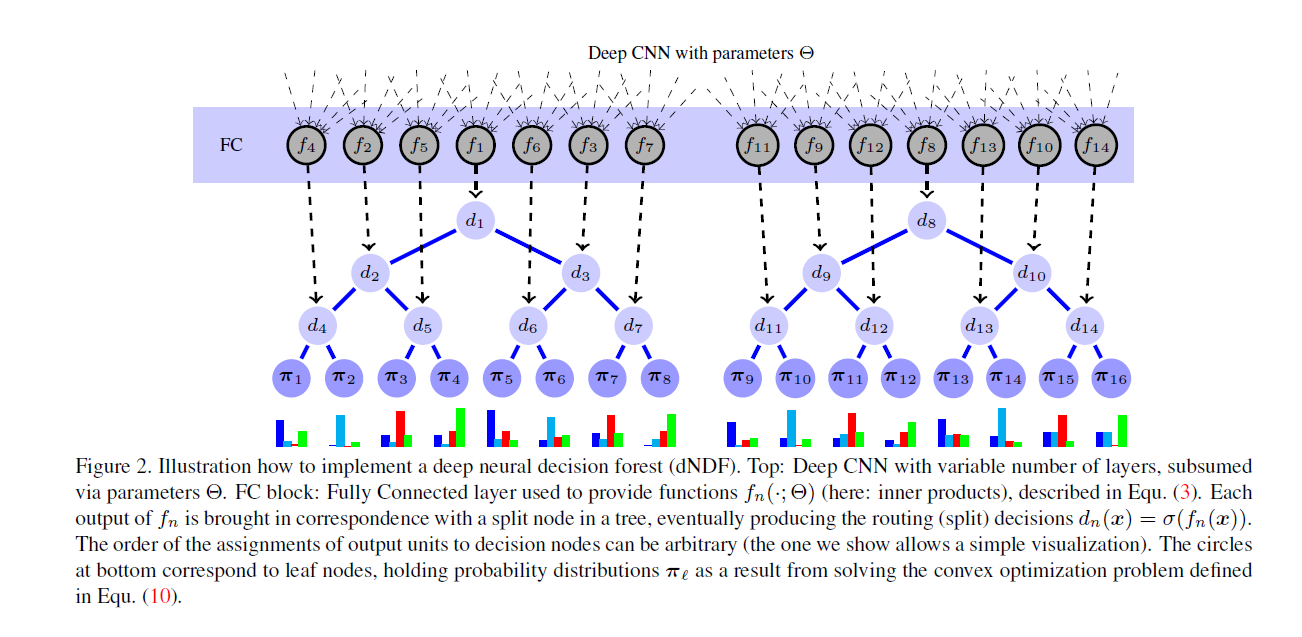}
\caption{The schematic illustration of decision nodes}  
\label{Fig.dNDF} 
\end{figure}

\subsubsection{TreeGrad}
\href{https://github.com/chappers/TreeGrad}{TreeGrad}, an extension of Deep Neural Decision Forests, 
reframes decision trees as a neural network through construction of three layers,
the decision node layer, routing layer and prediction (decision tree leaf) layer,
and introduce a stochastic and differentiable decision tree model.
In \href{https://arxiv.org/pdf/1904.11132.pdf}{TreeGrad}, we introduce a routing
matrix $Q$ which is a binary matrix which describes the relationship between the nodes  and the leaves. 
If there are $n$ nodes and  $\ell$ leaves, then $Q\in\{0,1\}^{\ell\times 2n}$,
where the rows of $Q$ represents the presence of each binary decision of the $n$ nodes for the corresponding leaf $\ell$.
We define the matrix containing the routing probability of all nodes to be $D(x;\Theta)$. 
We construct this so that for each node $j = 1,\cdots, n$, we concatenate each decision stump route probability
$$D(x;\Theta)=[d_{0_{+}}(x;\theta_0)\oplus\cdots \oplus d_{n_{+}}(x;\theta_n)\oplus d_{0_{-}}(x;\theta_0)\oplus\cdots \oplus d_{n_{-}}(x;\theta_n),$$
where $\oplus$ is the matrix concatenation operation,
and $d_{i_{+}}, d_{i_{-}}$ indicate the probability of moving to the positive route and negative route of node $i$ respectively.
We can now combine matrix $Q$ and $D(x;\Theta)$ to express $\mu_l$ as follows:
$$\mu_l=\exp(Q_{\ell}^T\log(D(x;\Theta)))$$
where $Q_{\ell}$ represents the binary vector for leaf $\ell$.
Accordingly, the final prediction for sample $x$ from the tree $T$ with decision nodes parameterized by $\Theta$ is given by
$$\mathbb{P}_{T}(y\mid x, \Theta, \pi)=\operatorname{softmax}(\pi^T\mu(x\mid \Theta, Q))$$
where $\pi$ represents the parameters denoting the leaf node values, 
and the row of $\mu(x\mid \Theta, Q)$ is the routing function which provides the probability that the sample $x$ will reach leaf $\ell$,
The formulation consists of three layers:
\begin{enumerate}
 \item decision node layer: $H_1=\tilde{W}x + \tilde{b}$;
 \item probability routing layer: $H_2=Q^T(\phi_2 \circ H_1)(x)$ where $\phi_2=(\log \circ \operatorname{softmax})(x)$;
 \item the leaf layer: $H_3=\pi^T(\phi_3 \circ H_2)(x)$ where $\phi_3=\exp(x)$.
\end{enumerate}
In short, this neural decision tree is expressed in as shown in \ref{Fig.3l}
\begin{equation}
T(x)=\pi^T\exp(Q^T(\log \circ \operatorname{softmax}) \circ (\tilde{W}x-\tilde{b})).
\end{equation}

\subsubsection{Neural tree ensembles}

\href{https://arxiv.org/pdf/1702.07360.pdf}{Neural Decision Trees} reformulate the random forest method of Breiman (2001) into a neural network setting

Neural Oblivious Decision Ensembles (NODE) is designed to work with any tabular data \cite{popov2019neural}. 
In a nutshell, the proposed NODE architecture generalizes ensembles of oblivious decision trees, 
but benefits from both end-to-end gradient-based optimization and the power of multi-layer hierarchical representation learning.
The NODE layer is composed of $m$ differentiable oblivious decision trees (ODTs) of equal depth $d$.
Then, the tree returns one of the $2^d$ possible responses, corresponding to the comparisons result.
The tree output is defined as 
$$h(x)=R[\mathbb{I}(f_1(x)-b_1),\cdots, \mathbb{I}(f_d(x)-b_d)]$$
where $\mathbb{I}(\cdot)$ denotes the Heaviside step function and $R$ is a d-dimensional tensor of responses.
We replace the splitting feature choice $f_i$ and the comparison operator $\mathbb{I}(f_1(x)-b_1)$ by their continuous counterparts.
The choice function $f_i$ is hence replaced by a weighted sum of features, with weights computed as entmax over the learnable
feature selection matrix:
  $$\hat{f}_i(x)=\sum_{j=1}^nx_j\operatorname{entmax}_{\alpha}(F_{ij}).$$
And we relax the Heaviside function $\mathbb{I}(f_i(d)-b_i)$  
as a two-class entmax $\sigma_{\alpha}(x)=\operatorname{entmax}_{\alpha}([0,x])$.
As different features can have different characteristic scales, we use the scaled version
$$c_i=\sigma_{\alpha}(\frac{f_i(x)-b_i}{\tau_i})$$
where $b_i$ and $\tau_i$ are learnable parameters for thresholds and scales respectively.
we define a ``choice” tensor $C$ the outer product of all $c_i$
$$C=\begin{pmatrix} & c_1 \\ & 1-c_1\end{pmatrix}\otimes\begin{pmatrix} & c_2 \\ & 1-c_2\end{pmatrix}
\otimes\cdots \otimes\begin{pmatrix} & c_d \\ & 1-c_d\end{pmatrix}.$$
The final prediction is then computed as a weighted linear combination of response tensor entries R
with weights from the entries of choice tensor $C$
$$\hat{h}(x)=\sum_{i_1,i_2,\cdots, i_d\in\{0,1\}^d}R_{i_1,\cdots, i_d}\cdot C_{i_1,\cdots, i_d}(x).$$
There is a supplementary code for the above method \UrlBigBreaks{https://github.com/Qwicen/node}.

\section{Neural decision trees}

Here neural decision trees refer to the models unite the two largely separate paradigms, deep neural networks and decision trees, 
in order to take the advantages of both  paradigms such as \cite{balestriero2017neural, yang2018deep}.

\subsection{Adaptive neural trees}

Deep neural networks and decision trees are united  via adaptive neural trees (ANTs) \cite{AdaptiveNeuralTrees19} that
incorporates representation learning into edges, routing functions and leaf nodes of a decision tree, 
along with a backpropagation-based training algorithm that adaptively grows the architecture from primitive modules (e.g., convolutional layers).

An ANT is defined as a pair $(\mathcal{T,O})$ where $\mathcal{T}$ defines the
model topology, and $\mathcal{O}$ denotes the set of operations on it.

An ANT is constructed based on three primitive modules of differentiable operations:
\begin{enumerate}
	\item Routers, $\mathcal{R}$: the router of each internal leaf sends samples from the incoming edge to either the left or right child.
	\item Transformers, $\mathcal{T}$: every edge of the tree has one or a composition of multiple transformer module(s).
	\item Solvers, $\mathcal{S}$: each leaf node operates on the transformed input data and outputs an estimate for the conditional distribution.
\end{enumerate}

Each input $x$ to the ANT stochastically traverses the tree based on decisions of routers 
and undergoes a sequence of transformations until it reaches a leaf node where the corresponding solver predicts the label $y$.

An ANT models the conditional distribution $p(y\mid x)$ as a hierarchical mixture of experts (HMEs) and
benefit from  lightweight inference via conditional computation.

Suppose we have $L$ leaf nodes, the full predictive distribution is given by
\begin{equation}\label{ANT}
p(y\mid x, \Theta)=\sum_{\ell}^{L}p(z_{\ell}=1\mid x, \theta,\psi) p(y\mid z_{\ell}=1, x, \phi,\psi)
\end{equation}
where $(\theta, \phi, \psi)$ summarize the parameters of router, transformer and solver modules in the tree. 
The mixing coefficient $\pi_{\ell}^{\theta,\psi}=p(z_{\ell}=1\mid x, \theta,\psi)$ quantifies the probability
that $x$ is assigned to leaf l and is given by a product of decision probabilities 
over all router modules on the unique path $\mathcal{P}_{\ell}$ from the root to leaf node $\ell$:
$${\pi}_{\ell}^{\theta,\psi}= \prod_{i\in \mathcal{P}_{\ell}} r_{j}^{\theta}(x_j^{\psi})^{\mathbb{I}(\ell\swarrow i)}(1 - r_{j}^{\theta}(x_j^{\psi}) )^{1-\mathbb{I}(\ell\swarrow i)}.$$
Here $r_{j}^{\theta}: \mathcal{X}_j\mapsto [0, 1]$, parametrized by $\theta$, is the  router of internal node $j$;
and  $\ell\swarrow i$ is a binary relation and is only true if leaf $\ell$ is in the left subtree of internal node $j$;
$x_j^{\psi}$ is the feature representation of $x$ at node $j$ defined as the result of composite transformation
$$x_j^{\psi}=(t^{\psi}_{e_n}\circ\cdots t^{\psi}_{e_2}\circ t^{\psi}_{e_1})(x)$$
where each transformer $t^{\psi}_{e_n}$ is a nonlinear function, parametrized by $\psi$, 
that transforms samples from the previous module and passes them to the next one.
The leaf-specific conditional distribution $p(y\mid z_{\ell}=1, x, \phi,\psi)$ is given
by its solver’s output $s_{\ell}^{\psi}(x_{parent(\ell)}^{\psi})$ parametrized by $\psi$. 

See its codes in \UrlBigBreaks{https://github.com/rtanno21609/AdaptiveNeuralTrees}.


\subsection{Tree ensemble layer}


The Tree Ensemble Layer (TEL)\cite{hazimeh2020tree}  is an additive model of differentiable decision trees.
The TEL is equipped  with a novel mechanism to perform conditional computation, during both training and inference,
by introducing a new sparse activation function for sample routing, along with specialized forward and backward propagation algorithms that exploit sparsity.

Assuming that the routing decision made at each internal node in the tree is independent of the other nodes,
the probability that $x$ reaches $\ell$ is given by:
\begin{equation}\label{soft routing}
P(x\to \ell)=\prod_{i\in A(\ell)} r_{\ell}(x, w_i)
\end{equation}
where $r_{\ell}(x, w_i)$ is the probability of node $i$ routing $x$ towards the subtree containing leaf $l$,
i.e., $$r_{\ell}(x, w_i)= \mathcal{S}(\left<x, w_i\right>)^{\mathbb{I}(\ell\swarrow i)}(1-\mathcal{S}(\left<x, w_i\right>))^{\mathbb{I}(\ell\searrow i)}.$$
Generally, the activation function $\mathcal{S}$ can be any smooth cumulant probability function.

For a sample $x \to\mathbb{R}^p$, we define the prediction of the tree as the expected value of the leaf outputs, i.e.,
\begin{equation}\label{TEL}
T(x)=\sum_{\ell\to L} P(x\to \ell) o_{\ell}
\end{equation}
where $\mathcal S$ in \eqref{soft routing} is so-called  smooth-step activation function, i.e.,
$$\mathcal S(t) = \left\{
\begin{array}{ll}
0, &t\leq -\frac{\gamma}{2},\\
\frac{2}{\gamma^3}t^3+\frac{3}{2\gamma}t+\frac{1}{2}, & -\frac{\gamma}{2}<t< \frac{\gamma}{2}\\
1,  &t\geq \frac{\gamma}{2}.
\end{array}
\right. $$

We can implement true conditional computation by developing specialized forward
and backward propagation algorithms that exploit sparsity.

See its implementation in \UrlBigBreaks{https://github.com/google-research/google-research}.

\subsection{Neural-backed decision trees}

Neural-Backed Decision Trees (NBDTs)\cite{nbdt} are proposed 
to leverage the powerful feature representation of convolutional neural networks
and the inherent interpretable of decision trees,
 which achieve neural network accuracy and require no architectural changes to a neural network.
Simply speaking, NBDTs use the convolutional network to extract the intrinsic representation 
and use the decision trees to generate the output based the learnt features.

The neural-backed decision tree (NBDT), has the exact same architecture as a standard neural network 
and  a subset of a fully-connected layer represents a node in the decision tree. 
The NBDT pipeline consists of four steps divided into a training phase and an inference phase. 
\begin{enumerate}
\item Build an induced hierarchy using the weights of a pre-trained network’s last fully-connected layer;
\item Fine-tune the model with a tree supervision loss;
\item For inference, featurize samples with the neural network backbone;
\item And run decision rules embedded in the fully-connected layer.
\end{enumerate}

The first step of training phase is to learn hierarchical decision procedure and 
the second step is to optimize the  pre-trained network and decision tree jointly.


Code and pretrained NBDTs can be found at \UrlBigBreaks{https://github.com/alvinwan/neural-backed-decision-trees}.

\section{Regularized decision trees and forests}
The regularization techniques are widely used in machine learning community to control the model complexity 
and overcome the over-fitting problem.
The regularization of nonlinear or nonparametric models is more difficult than the linear model.
There are two key factors to describe the complexity of decision trees: its depth and the total number of its leaves.
In univariate decision trees each intermediate node is a associated with  a single attribute.

We use the regularization techniques to select features such as \cite{deng2012feature, liu2014learning, raff2018fair}
and an interpretable model is learnt.
And the pruning techniques are used to find smaller models 
in the purpose to avoid over-fitting or deploy a lightweight models.


Like other iterative optimization methods, we take one training step to reduce the overall loss in boosted trees.
In another word, we add a new tree $f_K$ to update the model:
$$L(f_K)<L(f_{K-1}).$$ 
After each iteration, the model is more complicated and its cost is lower.
As a byproduct, it is prone to overfit.
In order to make a trade-off between bias and variance, 
the following regularized objective is to  minimize in \href{https://xgboost.readthedocs.io/en/latest/tutorials/model.html}{XGBoost} :
\begin{equation}\label{xGB training}
  L(f_K)=\sum_{i}\ell(\hat{y}_i, y_i)+\sum_{k=1}^K\Omega(f_k)
\end{equation}
where the complexity of tree is defined as
$\Omega(f) = \gamma T + \frac{1}{2}\lambda \sum_{j=1}^T w_j^2$
based on the  \eqref{chen} and $\hat{y}_i=\sum_{k=1}^K f_k(x_i)$.
Here $\ell(\hat{y}_i, y_i)$ is usually a differentiable convex loss function 
that measures the quality of prediction $\hat{y}_i$ on training data $(x_i, y_i)$
so that it is available to obtain the gradient $g_i=\frac{\partial \ell(\hat{y}_i, y_i)}{\partial \hat{y}_i}$
and the Hessian $h_i=\frac{\partial^2 \ell(\hat{y}_i, y_i)}{\partial^2 \hat{y}_i}$ when $\hat{y}_i=\sum_{k=1}^{K-1} f_k(x_i)$.
It is in alternating approach to train a new tree.
First, we fix the number of leaves $T$, we can reduce \ref{xGB training} via the surrogate loss
$$\arg\min_{w} \sum_{i}\underbrace{\ell(\hat{y}_i, y_i)+f_K(x_i)g_i+\frac{1}{2}f_K(x_i)^2h_i}_{\text{taking Taylor expansion at $\hat{y}_i=\sum_{k=1}^{K-1} f_k(x_i)$} }
+\gamma T + \frac{1}{2}\lambda \sum_{j=1}^T w_j^2+\sum_{k=1}^{K-1}\Omega(f_k)$$ 
i.e., 
\begin{equation}
  \begin{split}w_j^\ast &= -\frac{G_j}{H_j+\lambda}\\
  \text{obj}^\ast &= -\frac{1}{2} \sum_{j=1}^T \frac{G_j^2}{H_j+\lambda} + \gamma T\end{split}
\end{equation}
where $G_j=\sum_{q_K(x_i)=j} g_i$ and $H_j=\sum_{q_K(x_i)=j} h_i$.
Second, we will split a leaf into two leaves if the following gains is positive
$$Gain = \frac{1}{2} \left[\frac{G_L^2}{H_L+\lambda}+\frac{G_R^2}{H_R+\lambda}-\frac{(G_L+G_R)^2}{H_L+H_R+\lambda}\right] - \gamma.$$
And it is equivalent to the pruning techniques in tree based models. 

This regularization is suitable for the univariate decision trees. 
And the training procedure is still greedy.
In the following, we will review some regularization techniques in tree-based methods.

\subsection{Regularized decision trees}

\subsubsection{Regularized soft trees}

We can directly apply the norm penalty to train soft trees \cite{irsoy2012soft} in differentiable trees and neural trees as
we apply norm penalty to deep learning.
For example, Olcay Taner Yıldız and Ethem Alpaydın introduce local
dimension reduction via $L_1$ and $L_2$ regularization for feature selection and smoother
fitting in \cite{yildiz2013regularizing}.
And we can induce the sparse \href{https://www.aaai.org/ojs/index.php/AAAI/article/view/4505/4383}{weighted oblique decision trees} 
as shown in \href{https://www.aaai.org/ojs/index.php/AAAI/article/view/4505/4383}{SWOT}.
Another aim of sparse decision trees is to regularize with sparsity for interpretability.

\subsubsection{Sparse decision trees}

Sparse decision trees are oblique or multivariate decision trees regularized by sparsity-induced norms to avoid over-fitting.

We can generate sparse trees in optimal trees such as \cite{hu2019optimal, blanquero2019sparsity, lin2020generalized}.
The optimal decision trees can learn the decision template from the data with theoretical guarantee.

The OSDT \cite{hu2019optimal}, proposed by Xiyang Hu, Cynthia Rudin and Margo Seltzer
is desiigned for binary features. 

Generalized and Scalable Optimal Sparse Decision Trees(GOSDT) provides a general framework for decision
tree optimization that addresses the two significant open problems in the area: treatment of imbalanced data and fully optimizing over continuous variables.

And it is official implementation in 
\UrlBigBreaks{https://github.com/xiyanghu/OSDT} and
\UrlBigBreaks{https://github.com/Jimmy-Lin/GeneralizedOptimalSparseDecisionTrees}.

\subsection{Regularized additive trees}
Regularized decision forests is not only the sum of regularized decision trees.
For example, Heping Zhang and Minghui Wang in \cite{zhang2009search} propose a specific method to find a sub-forest 
(e.g., in a single digit number of trees) that can achieve the prediction accuracy of a large random forest (in the order of thousands of trees).

The complexity of the additive trees are the sum of the single tree complexity.
At one hand, we regularize the objective functions in order to control the complexity oof the addtively trained trees as in xGboost.
At another hand, we want to regularize the whole additive trees in order to keep the balance between the bias and variances.

\subsubsection{Regularized random forests}
Regularized random forests are used for feature selection specially in gene selection 
\cite{deng2013gene, liu2014learning, uzdilli2015swiss-chocolate, raff2018fair}.
And there is its open implementation at \UrlBigBreaks{https://cran.r-project.org/web/packages/RRF/index.html}


Sparse Projection Oblique Randomer Forests (SPORF)\cite{JMLR:v21:18-664} is yet another decision forest
which  recursively split along very sparse random projections. 
SPORF uses very sparse random projections, i.e., linear combinations of a small subset of features
Its official web is \UrlBigBreaks{https://neurodata.io/sporf/}.

\subsubsection{Regularized Boosted Trees}
We can construct the regularzied boosting machines based on trees such as \cite{johnson2013learning, chen2015xgboost, chen2014higgs}.

In \UrlBigBreaks{https://arxiv.org/abs/1806.09762}, we regularize gradient boosted trees by introducing subsampling and employ a modified shrinkage algorithm so that at every boosting stage the estimate is given by an average of trees. 
\section{Conditional computation}

Conditional Computation refers to a class of algorithms in which each input sample uses a different part of the model, 
such that on average the compute, latency or power (depending on our objective) is reduced.
To quote Bengio et. al
\begin{quote}
Conditional computation refers to activating only some of the units in a network, in an input-dependent fashion.
 For example, if we think we’re looking at a car, we only need to compute the activations of the vehicle detecting units, 
 not of all features that a network could possible compute. 
 The immediate effect of activating fewer units is that propagating information through the network will be faster, 
 both at training as well as at test time. 
 However, one needs to be able to decide in an intelligent fashion which units to turn on and off, depending on the input data. 
 This is typically achieved with some form of gating structure, learned in parallel with the original network.
\end{quote}

Another natural property in trees is conditional computation, 
which refers to their ability to route each sample through a small number of nodes (specifically, a single root-to-leaf path).
Conditional computation can be broadly defined as the ability of a model to activate only a small part of its architecture
in an input-dependent fashion.

Conditional Networks \cite{ioannou2016decision} is a fusion of conditional computation with representation learning 
and achieve a continuum of hybrid models with different ratios of accuracy vs. efficiency.

\subsection{Unbiased recursive partitioning}
Wei-Yin Loh and Yu-Shan Shih present an algorithm called
QUEST that has negligible bias in \cite{loh1997split}.

A unified framework for recursive partitioning is proposed in \cite{hothorn2006unbiased, schlosser2019the},
which embeds tree-structured regression models into a well defined theory of conditional inference procedures.

We focus on regression models describing the conditional distribution of a response variable $Y$ 
given the status of $m$ covariates by means of tree-structured recursive partitioning.
The response $Y$ from some sample space $\mathcal{Y}$ may be multivariate as well. 
The $m$ covariates $X = (X_1, \cdots ,X_m)$ are element of a sample space $\mathcal{X} = \mathcal{X}_1 \times \cdots\mathcal{X}_m$.
We assume that the conditional distribution $D(Y\mid X)$ of the response $Y$ 
given the covariates $X$ depends on a function $f$ of the covariates
$$D(Y\mid X)=D(y\mid X_1, \cdots ,X_m)=D(y\mid f(X_1, \cdots ,X_m))$$
where we restrict ourselves to partition based regression relationships.
A regression model of the relationship is to be fitted based on a learning sample $\{Y, X_1, \cdots ,X_m\}$.

The following generic algorithm implements unbiased recursive binary partitioning:
\begin{enumerate}
  \item For case weights $w$ test the global null hypothesis of independence 
  between any of the $m$ covariates $X$ and the response $Y$.
	Stop if this hypothesis can not be rejected. 
	Otherwise select the $j^{\ast}$th covariate $X_{j^{\ast}}$ with strongest association to $Y$.
	\item Choose a set $A^* \subset X$  in order to split $X_{j^{\ast}}$ into two disjoint sets $A^*$ and $A^*-X_{j^{\ast}}$.
	\item Repeat recursively steps 1 and 2.
\end{enumerate}

Each node of a tree is represented by a vector of case weights having non-zero elements 
when the corresponding observations are element of the node and are zero otherwise. 

We need to decide whether there is any information about the response variable covered by any of the $m$ covariates.
The fundamental problem of exhaustive search procedures have been known for a long time 
is a selection bias towards covariates with many possible splits or missing values.

Here are the R packages on recursive partitioning 

\UrlBigBreaks{https://www.rdocumentation.org/packages/partykit/versions/1.2-11}. 

\subsection{Bonzai}

\href{http://proceedings.mlr.press/v70/kumar17a.html}{Ashish Kumar, Saurabh Goyal, and Manik Varma} develop a tree-based algorithm called `Bonzai'
\begin{equation}\label{Bonzai}
T(x)=\sum_{k}\mathbb{I}_k(x)W_k^TZx\circ \tanh(\sigma V_k^TZx)
\end{equation}
where $\circ$ denotes the element-wise Hadamard product, 
$\sigma$ is a user tunable hyper-parameter, 
$Z$ is a sparse projection matrix 
and Bonsai’s tree is parameterized by $\mathbb{I}_k$ , $W_k$ and $V_k$
where $\mathbb{I}_k(x)$ is an indicator function taking the value 1 if
node $k$ lies along the path traversed by $x$ and 0 otherwise
and $W_k$ and $V_k$ are sparse predictors learnt at node $k$. 
Bonsai computes $\mathbb{I}_k$ by learning a sparse vector $\theta$ at each internal node such
that the sign of $\theta^T Zx$ determines whether point $x$ should be branched to the node’s left or right child.
In fact, the indicator function is relaxed when implemented.
A gradient descent based algorithm with iterative hard threshold (IHT) was found to solve the optimization of Bonzai.

See its implementation in  \UrlBigBreaks{https://github.com/Microsoft/EdgeML}.

\section{Discussion}

Decision trees for regression or classification takes diverse forms as shown as above.
We study this field from different perspectives: ensemble methods, Bayesian statistics, adaptive computation and conditional computation. 
It is a theoretical overview on tree-based models associated with some implementation of new methods.
Decision tree is a fast developing field and interactive with diverse fields.
It seems simple and intuitive while powerful and insightful.

In the end, we identify some trend for future research.
\begin{itemize}
  \item sufficient and compact representation of decision trees;
  \item combination of decision trees and deep learning;
  \item interpretation of tree-based models.
\end{itemize}

\bibliographystyle{plain}
\bibliography{Note}

\end{document}